\documentclass[conference]{IEEEtran}
\IEEEoverridecommandlockouts
\usepackage{cite}
\usepackage{amsmath,amssymb}
\usepackage{algorithmic}
\usepackage{graphicx}
\usepackage{textcomp}
\usepackage{subfig}
\usepackage{booktabs}
\usepackage{multirow}
\usepackage{comment}
\usepackage{xcolor}
\usepackage [autostyle, english = american]{csquotes}
\MakeOuterQuote{"}
\def\BibTeX{{\rm B\kern-.05em{\sc i\kern-.025em b}\kern-.08em
    T\kern-.1667em\lower.7ex\hbox{E}\kern-.125emX}}
\begin{document}

\title{Learning Occupancy Function from Point Clouds for Surface Reconstruction\\
}

\author{\IEEEauthorblockN{Meng Jia}
\IEEEauthorblockA{\textit{Electrical Engineering and Computer Science} \\
\textit{York University}\\
Toronto, Canada \\
mjia@eecs.yorku.ca}
\and
\IEEEauthorblockN{Matthew Kyan}
\IEEEauthorblockA{\textit{Electrical Engineering and Computer Science} \\
\textit{York University}\\
Toronto, Canada \\
mkyan@eecs.yorku.ca}
}

\maketitle

\begin{abstract}
Implicit function based surface reconstruction has been studied for a long time to recover 3D shapes from point clouds sampled from surfaces. Recently, Signed Distance Functions (SDFs) and Occupany Functions are adopted in learning-based shape reconstruction methods as implicit 3D shape representation. This paper proposes a novel method for learning occupancy functions from sparse point clouds and achieves better performance on challenging surface reconstruction tasks. Unlike the previous methods, which predict point occupancy with fully-connected multi-layer networks, we adapt the point cloud deep learning architecture, Point Convolution Neural Network (PCNN), to build our learning model. Specifically, we create a sampling operator and insert it into PCNN to continuously sample the feature space at the points where occupancy states need to be predicted. This method natively obtains point cloud data's geometric nature, and it's invariant to point permutation. Our occupancy function learning can be easily fit into procedures of point cloud up-sampling and surface reconstruction. Our experiments show state-of-the-art performance for reconstructing With ShapeNet \cite{chang2015shapenet} dataset and demonstrate this method's well-generalization by testing it with McGill 3D dataset \cite{siddiqi2008retrieving}. Moreover, we find the learned occupancy function is relatively more rotation invariant than previous shape learning methods.
\end{abstract}


\section{Introduction}
Motivated by the success of deep learning in image processing and analysis, learning-based 3D shape processing has received considerable attention. Leveraging the rich shape priors learned from data, these methods may work on challenging scenarios that traditional methods cannot handle. Different types of shape representations are used in those approaches: 3D volumetric grids (array of voxels), point clouds, triangle meshes, multi-view images or depth maps, etc. Among these efforts, point cloud deep learning methods attempt to generalize the techniques devised for Convolutional Neural Networks (CNNs) \emph{directly} to point clouds (which is the most available and flexible form of 3D shape representation), providing versatile and general learning frameworks. These methods apply end-to-end deep learning to point clouds, mainly focusing on discriminative tasks like point cloud classification/recognition and part/semantic segmentation.

\begin{figure}
	\includegraphics[width=\linewidth]{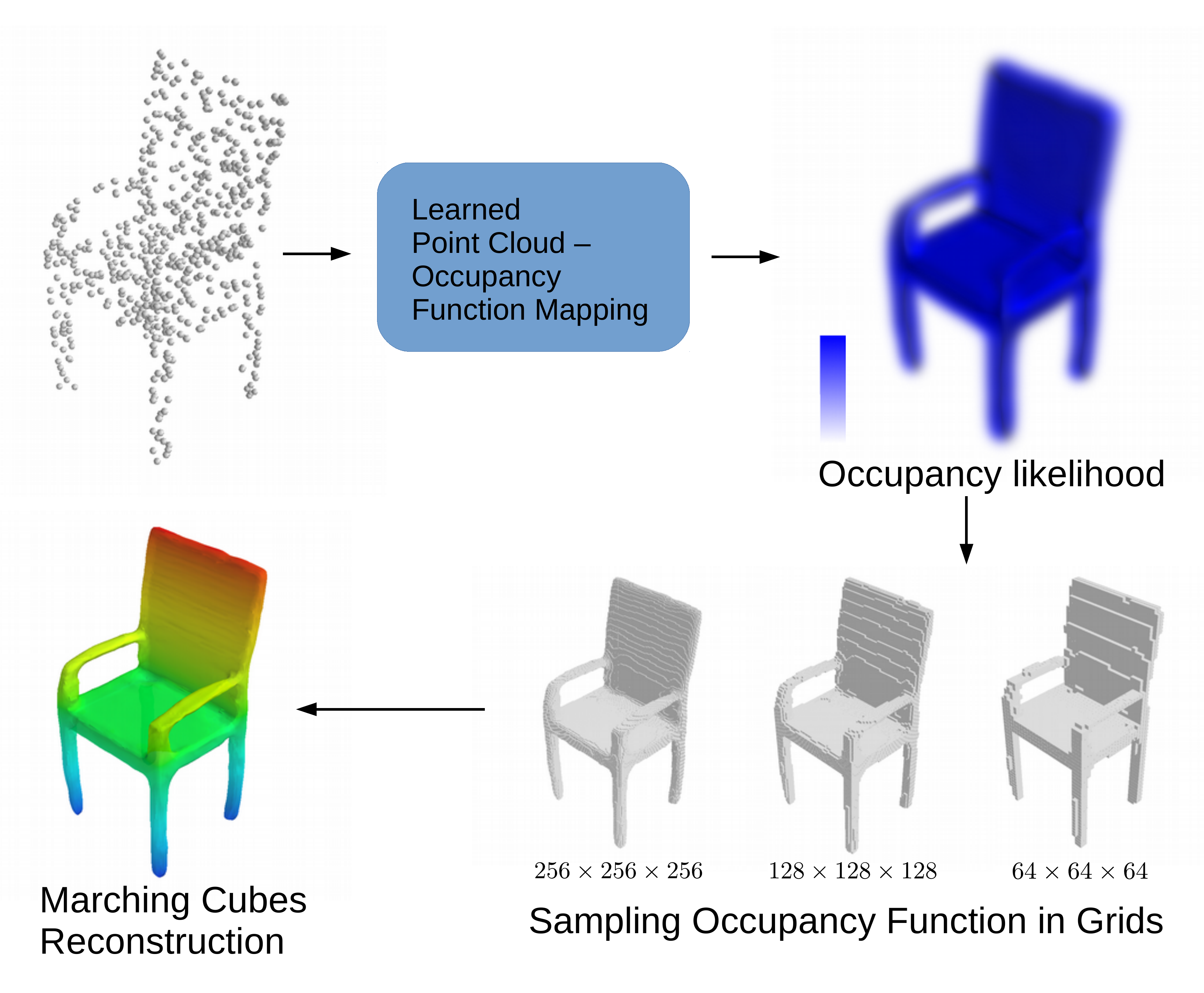}
	\caption{Illustration of the proposed method, shown as a pipeline from the input point cloud to reconstructed mesh.}
	\label{fig_system}
\end{figure}

Such point could deep learning has facilitated other studies like 3D shape representation, generation, point cloud up-sampling, and surface reconstruction. Those tasks are related and sometimes interchangeable. Most of these tasks come down to learning efficient representations/embedding of point clouds and mappings between them. Throughout these varied methods, we are especially interested in the approaches that learn implicit functions\footnote{Implicit funtions including Signed Distance Functions(SDFs), partial derivative fields, or indicator function. These function represent shapes as continuous volumetric function. In this paper, we will treat \emph{indicator function} and \emph{occupancy function} as synonymous and uses them interchangeably according to context.}: Signed Distance Function (SDF) or Occupancy/Indicator Function. The advantages of this approach are three-fold. First, learning 3D volumetric indicator function or signed distance field is native for neural networks (as universal function approximators). Second, from this representation, we can easily transform the generated shape into meshes (which will be manifold and watertight), voxel grids (by sampling over a grid to form query points), point clouds, and parametric representations (through fitting to spline or volumetric primitives). Third, such representations drastically simplify shape manipulations like surface filtering, morphological, and boolean operations, as we can conduct corresponding arithmetic operations directly on their volumetric indicator functions.

However, recent implicit function-based shape-learning methods focus only more on shape generation \cite{chen2019learning}. They are usually incapable of capturing local geometry details \cite{mescheder2018occupancy}. In this paper, we propose a novel method that adapts Point Convolution Neural Network (PCNN) \cite{atzmon2018point}, a recent end-to-end learning architecture first applied to discriminative tasks, to learn continuous 3D functions. With this method, the network can produce occupancy function from sparse point cloud input, and we implement a surface reconstruction system based on this occupancy network. 

Our method is sketched in Figure \ref{fig_system} and detailed in Section \ref{Sec_method}. The proposed PCNN-based occupancy network is treated as a frontend, and we use the Marching Cubes algorithm \cite{lewiner2003efficient} to extract an iso-surface as the reconstructed structure. The network takes a point cloud as input and maps the shape represented by this point cloud to its indicator function. The predicted indicator function is then voxelized into three grids (sampled coarse-to-fine hierarchically). Lastly, we run Marching Cubes to produce the final triangle mesh.

Experiments conducted on the ShapeNet dataset \cite{chang2015shapenet} indicate that our method outperforms other learning-based methods \cite{choy20163d, liao2018deep, mescheder2018occupancy}. Previous methods implement an encoder-decoder framework and thus can only generate stereotypical shapes of each category of objects. In contrast, by integrating shape features in the geometry domain and relying more on local shape information carried by point clouds, our method may capture and recover unique shape details. That also gives it better generalization ability and achieves a certain degree of rotation invariant like traditional surface reconstruction methods. 

\section{Related Works}

\subsection{Classic Approaches of Surface Reconstruction} \label{3DRFPC}

There has been a vast amount of effort devoted to the study of 3D surface reconstruction in the past several decades. These approaches work with different surface representations, reconstructing parametric surfaces \cite{rekanos2008shape, gao2019deepspline, groueix2018atlasnet}, polygon meshes \cite{amenta2002simple, kolluri2004spectral}, or implicit functions of the surface \cite{carr2001reconstruction, kazhdan2013screened}. The last category above is most relevant to our work.

These methods produce as reconstructed shape an iso-surface or level-set of a volumetric representation of shapes, either Signed Distance Fields (SDFs), partial derivative fields, or indicator functions. Early works \cite{hoppe1992surface, boissonnat2002smooth} fit the tangent plane locally to compute a distance field and triangulating its zero-set. Since they require a good estimation of the tangent planes, if the point cloud is sparsely or non-uniformly sampled, incomplete, noisy, or misaligned, their performance is poor. Moving Least Square (MLS) methods \cite{fleishman2005robust, cheng2008survey} approximate the shape with spatial low-degree polynomial and recover SDFs with least square optimization. Other approaches \cite{kazhdan2005reconstruction, kazhdan2006poisson, kazhdan2013screened} formulate it as a volumetric shape estimation problem. In this way, volume smoothness prior (other than surface smoothness prior) is imposed and results in better robustness. Poisson surface reconstructions recover implicit function from \emph{oriented} points (i.e., location and surface normal tuples). Oriented points are viewed as a discrete sampling of indicator functions' gradient field. \cite{kazhdan2005reconstruction} solves this Poisson problem by transforming it into the Fourier domain, computing Fourier coefficients of the indicator function, and \cite{kazhdan2006poisson} solve it directly in a coarse-to-fine hierarchical manner. Poisson methods require a good surface normal estimation beforehand.

Among the traditional methods, the RBFs based methods \cite{carr2001reconstruction} could be viewed as a hand-crafted version of our method. Carr et al. \cite{carr2001reconstruction} use overlapped RBFs to approximate SDFs. RBFs centers are scattered around the input points (on-surface) and along the direction of local surface normals (off-surface). Coefficients are associated with each RBF, and they are optimized to fit input points to approximate SDFs. In our occupancy function learning method, the learned feature fields are also composed of RBFs. The counterparts of the off-surface RBF centers are the elements in the RBF filter kernels. The RBFs centers are scatted in all directions and have varied scales and offsets in different PCNN layer. RBFs are weighted by the hidden nodes' outputs, non-linearly depending on the network's parameters (learned shape prior) and the input point cloud (one instance of observation).

\subsection{Point Cloud Deep Learning and Shape Representation}

\paragraph{Point Cloud Deep Learning for Recognition and Classification} Recently, applying deep learning techniques to point clouds has received significant attention. The early work is stimulated by the advances of 3D sensing and the availability of point cloud data. PointNet \cite{qi2017pointnet} is a pioneer work in this field. By using symmetric functions (mean- and max-pooling) and point-wise non-linear mapping for points feature learning, PointNet is computationally efficient and permutation invariant. 
Deep-RBFNet \cite{chen2018deep} is inspired by RBF-based surface representation \cite{carr2001reconstruction} and Radial Basis Function Network \cite{broomhead1988radial}. Before feeding into a PointNet-like architecture, they propose first encoding each input 3D point with RBF kernels scattered in the space whose center and width are trainable parameters. \cite{ben20183dmfv} use a Fisher Vector \cite{sanchez2013image} to be the non-linear activation. Since the Fisher Vector is based on a Gaussian Mixture Model, it is analogous to RBF encoding \cite{chen2018deep}.

A considerable amount of methods leverage the point-wise local or global structure, extracting nearest neighbors hierarchical graphs to construct efficient learning architectures. The structure could be extracted from either bottom-up \cite{li2018pointcnn, wang2018dynamic, shen2018mining, hua2018pointwise, liu2019relation} or top-down \cite{klokov2017escape, li2018so, lei2019octree} direction. PointConv \cite{wu2019pointconv} and PCNN \cite{atzmon2018point} aim to extend the image discrete convolution operation to point clouds. The key is transforming discrete points back to continuous function over the 3D space. PointConv \cite{wu2019pointconv} solve this problem with a kernel-based point density estimation, and the point cloud shape is approximated as its density field. PCNN \cite{atzmon2018point} propose to do this transform by directly place a fixed-size RBF at each point and sum these RBFs over all of the points, producing a smooth field that contours the distribution of points. The method \cite{atzmon2018point} will be briefly reviewed in section \ref{sec_OLPCCO} when presenting our occupancy learning method.

\paragraph{Learning Approach of Shape Representations and Reconstruction}
Deep learning techniques have facilitated studies of other point cloud tasks. In shape representation learning and reconstruction, learn-based methods use data-driven prior (other than smooth prior), making them able to work on extremely sparse and noisy point clouds which traditional methods cannot handle.

Fan et al. \cite{fan2017point} introduced a 3D object reconstruction method that generates point cloud from a single image. Achlioptas et al. \cite{achlioptas2017learning} use Auto-Encode (to learn a latent code of shapes) and GANs \cite{goodfellow2014generative} (to generate latent code for shape decoder) to generate point clouds 3D objects. Some other methods use a deformable mesh \cite{wang2018pixel2mesh} or a 2D grid of points \cite{yang2018foldingnet} to represent 3D shapes, training a model to learn the geometric deformation. AtlasNet \cite{groueix2018atlasnet} propose to represent 3D shapes with a collection of parametric surface elements (Atlas indicates a set of small squares fitting locally to the surface), and train shape Auto-Encoder and reconstruction 3D shape with Atlas.

Point cloud consolidation or up-sampling \cite{yifan2019patch, yu2018pu, yu2018ec, roveri2018pointpronets} are also explored from a machine learning perspective. In a traditional 3D shape digitization pipeline, consolidation is considered as a crucial preprocessing step before conducting surface reconstruction \cite{huang2009consolidation}. Compared with these works, our method is more general and versatile as it represents 3D shapes explicitly and completely. 

Implicit surface representation is recently adopted in learning-based shape reconstruction methods \cite{huang2009consolidation, mescheder2018occupancy, park2019deepsdf, chen2019learning, atzmon2019controlling}. These methods aim to train a deep neural network modeling SDF or indicator function of a 3D shape. Most of those networks are composed of an point cloud shape encoder and a decoder pair. The decoder transforms the encoder's shape abstract to generate a function in 3D space $C\left(\mathbb{R}^{3},\mathbb{R}^{J}\right)$, either SDFs (point-wise regressive model) or indicator functions (point-wise binary classifier). Our method's key distinction is that we use PCNN \cite{atzmon2018point} to build a unified network without encoder or decoder, which is better at preserving local shape details and achieves rotation invariant to a certain extent.

\section{Method} \label{Sec_method}

In this section, we first formulate the occupancy function learning as a per-point classification problem. We then describe how we solve this occupancy learning problem by adapting the point cloud convolutional operation proposed in \cite{atzmon2018point}. The way to generate training data is discussed and compared with Occupancy Network \cite{mescheder2018occupancy}. Lastly, we show the way of reconstructing mesh from the learned occupancy function.

\begin{figure}[t]
	\includegraphics[width=\linewidth]{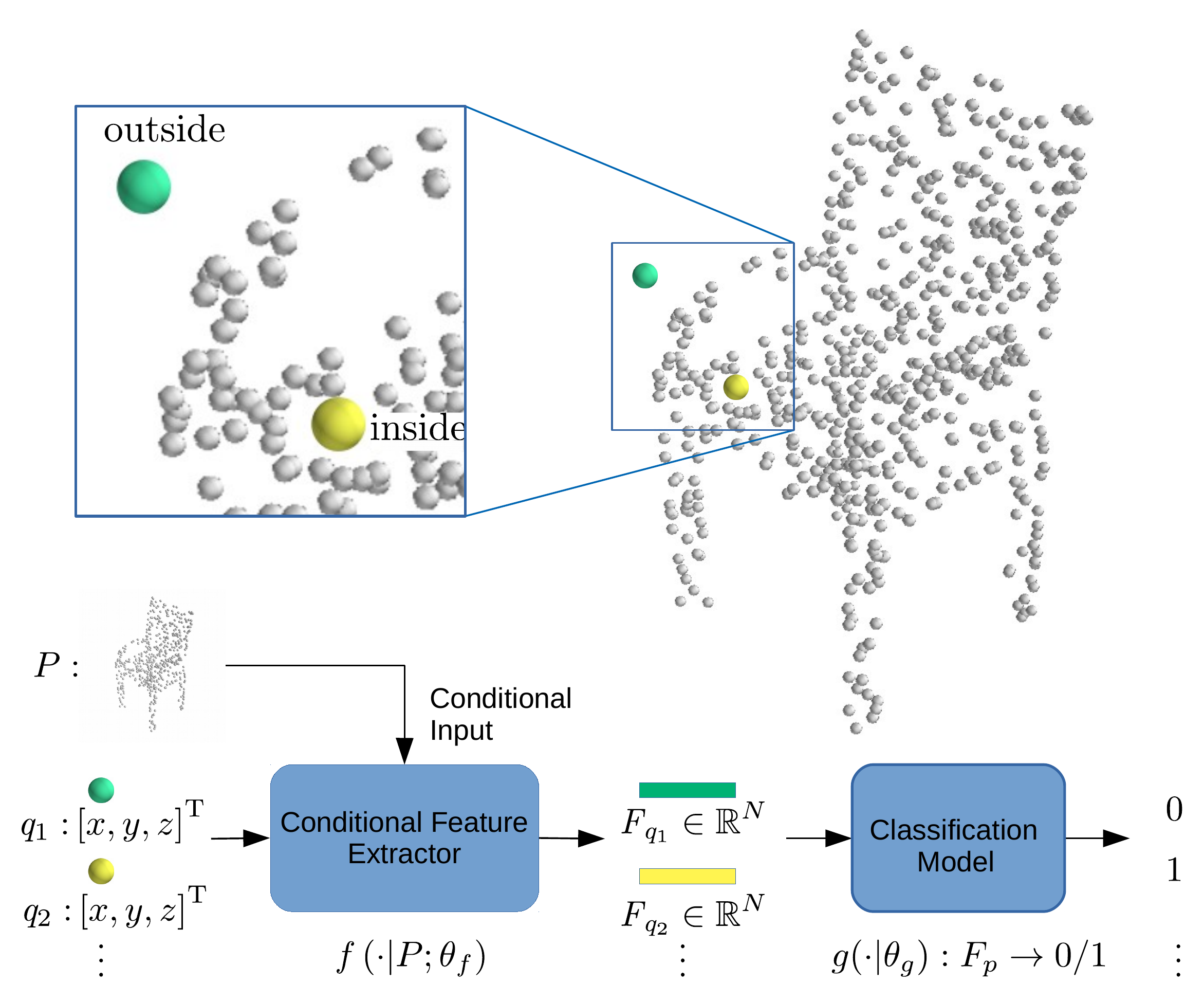}
	\caption{High level abstract and notations of our classifier.}
	\label{fig_sys_abstract}
\end{figure}

\subsection{Occupancy Function Learning} \label{OFL_SUBSEC}
\textbf{Modeling occupancy function with a binary classifier} Occupancy function is a two-valued function in 3D space, $\mathit{o:}\:\mathbb{R^{\textrm{3}}}\rightarrow\left\{ 0,1\right\} $, indicating whether this point is an occupant of a solid $M$ (interior of the object) or not (exterior), i.e., $\boldsymbol{\mathbf{1}}\left(q\in M\right)$, and it's therefore called indicator function in computer graphic context. $\partial M$ indicates the surface of solid $M$, and it is what we want to recover from an observed point cloud $P:\left\{ {p}_{1},{p}_{2},\ldots {p}_{n}\right\} $, where ${p}_{n}\in \mathbb{R}^{3}$. Following \cite{mescheder2018occupancy}, we formulate occupancy learning as a 3D point classification problem, learning a network $g\left(q\mid {P};\theta\right)$ to approximate $o\left(q;{P}\right)$, i.e., for a query point $q \in Q$, its likelihood of interior or exterior is predicted as the network's output.

Since this is a binary classification problem, we could easily form it as the model shown in Figure \ref{fig_sys_abstract}. Those  query points $q\in {\mathbb{R}}^{3}$ are first fed to a conditional feature function $f:{\mathbb{R}}^{3} \rightarrow {\mathbb{R}}^{N}$, and then classified. We use a fully connected Relu network with a softmax layer as the classifier:
\begin{equation} \label{equ_classifier}
g\left(q|P;\left\{ \theta_{f},\theta_{g}\right\} \right)=\mathrm{softmax}\left[\mathrm{CF_{\theta_{g}}}\left(f_{\theta_{f}}\left(q|P;\right)\right)\right].
\end{equation}
Note that outputs of those individual query points conditioned only on the input point cloud $P$ represent the shape we reconstruct and are independent of each other.

Training of this entire model is done by minimizing the loss function:
\begin{equation} \label{occu_learning_loss}
\small
\mathcal{L_{B}}\left(\theta\right)=\frac{1}{|\mathcal{B}|}\sum_{i=1}^{|\mathcal{B}|}\sum_{j=1}^{\mathcal{K}}\mathcal{L}\left(g\left(q_{j}^{(i)}\mid P_{i};\theta\right),\mathit{o\left(q_{j}^{(i)};P_{i}\right)}\right),
\end{equation}
where $q_{j}^{(i)}$ denotes the $j$th random query point corresponding to the $i$th point cloud in a batch $\mathcal{B}$. We use cross-entropy as the loss function $\mathcal{L}\left(\cdot\right)$.

\textbf{Training data} Query points $q$ could potentially be any point in the 3D space. Training data of this network is a number of point clouds representing shapes corpus. Each point cloud (on-surface points) is associate with a set of off-surface points (queries during training) sampled and labeled inside and outside of the surface defined by the point cloud. This form of data is not like any sensor's raw data and thus does not naturally exist. However, there are many 3D shape datasets contain thousands of high-quality 3D mesh models. Given a watertight (so that there is no ambiguity of interior and exterior) 3D mesh, we can randomly take an arbitrary amount of points (queries) around it and make Ray-test for each query points to determine whether it's inside or outside (ground-truth label) of the mesh. The input point clouds could be sampled from the mesh pure randomly, like drawing for uniform distribution, or deliberately, like simulating partially viewed point clouds. In our study and experiments, we take a subset from ShapeNet \cite{chang2015shapenet}, a multi-category 3D mesh dataset, and use the method proposed in \cite{stutz2018learning} to make the meshes watertight. Point clouds $P$ and query points $Q$ are sampled from the surface and space (in/out) around it.

\begin{figure}[t!]
	\includegraphics[angle=90,width=\linewidth]{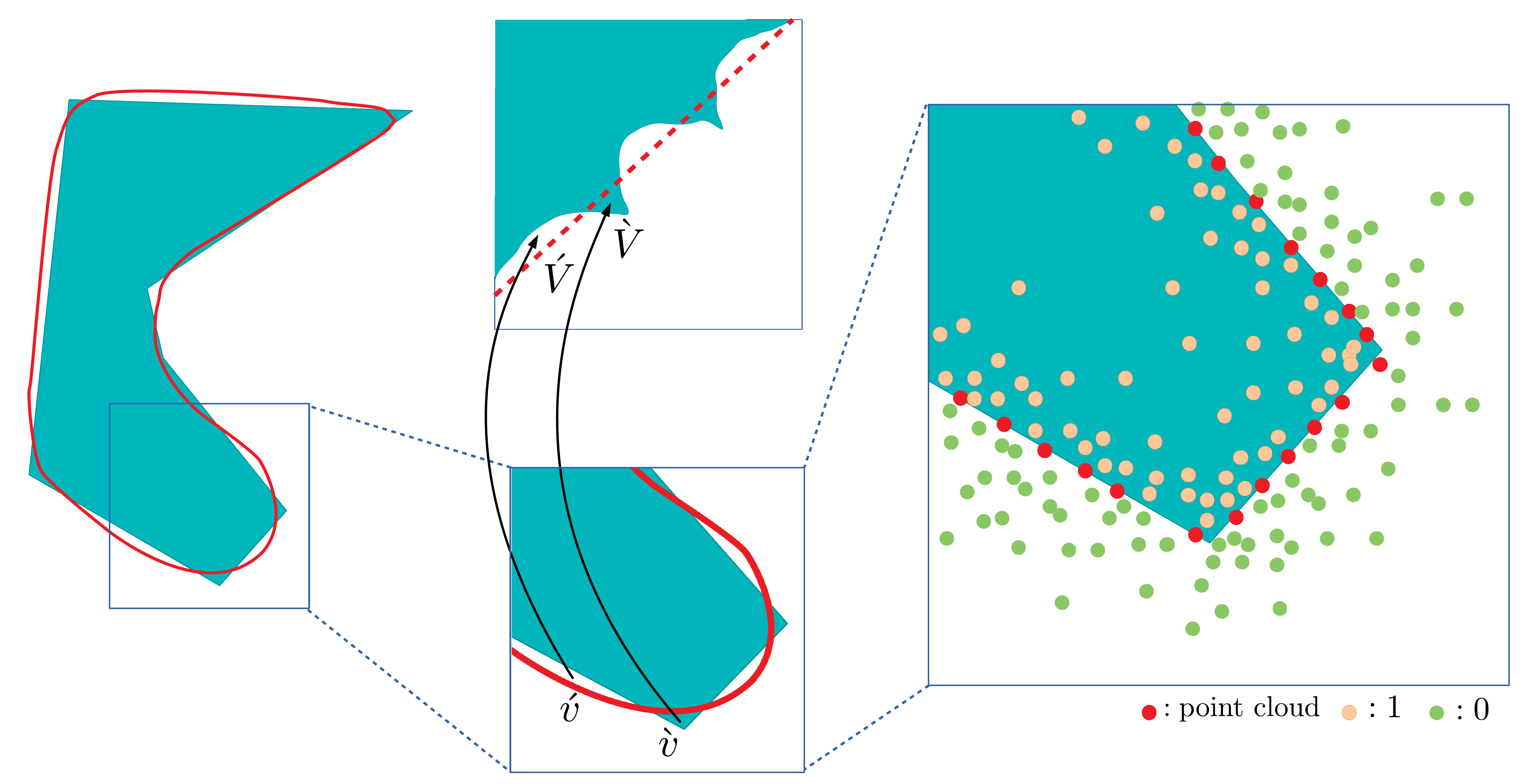}
	\caption{2D illustration of the classifier's decision boundary (in 2D and feature space) and the way of generating training data. On the left side, a target shape (cyan area) is shown with an error-containing decision boundary (red curve) representing the learned occupancy function. Each point is mapped to feature space and classified. The red dotted line indicates the decision boundary of the softmax classifier. The right figure shows how the individual training query points are generated from a particular shape.}
	\label{fig_3d_and_feature_space}
\end{figure}

Theoretically, an infinite number of random query points could be sampled around the watertight mesh. However, the way of sampling those points matters for learning unbiased occupancy function. The classification model (\ref{equ_classifier}) is not geometric aware. Any query points fed to it are treated as individual entries. Suppose the sample points distribute different density inside and outside of $M$. In that case, we could not expect the network to learn an unbiased surface around $\partial M$. An illustration of this issue is shown in Figure \ref{fig_3d_and_feature_space}. A possible solution to this problem is to sample the bounding box of the objects uniformly. The drawback of this solution is obvious: points locate far from the surface (inside and outside) are easy to classify, and error happens more around the surface $\partial M$. Uniformly sampling the bounding box will make network training less efficient. As shown in Figure \ref{fig_3d_and_feature_space}, if training converges, areas of positive error $\acute{v}$ and negative error $\grave{v}$ contribute most to the loss function (\ref{occu_learning_loss}). In feature space, the corresponding area $\acute{V}$ and $\grave{V}$ push the decision boundary towards different directions. Imprecisely, if the negative and positive samples are symmetrically distributed around the local mesh surface, the decision boundary will be overlapped with it without bias. So, we first sample points from the object surface, add a small offset to move them off-surface and label them correspondingly. The offsets are draw from zero-mean normal distributions. See Figure \ref{fig_3d_and_feature_space} (right) for a demonstration (in 2D) of this process's result. As most of nature and human-made objects consist of more convex curvature surface than concave, the way will over-sample outside (some small object parts could be narrower than the width of offset's distribution). Ideally, the Monte-Carlo method can be used to estimate the sample and re-balance it during training. However, we don't involve that in our experiments and empirically find that it isn't a serious problem. 

\subsection{Occupancy Learning with Point Cloud Convolutional Operation} \label{sec_OLPCCO}
This section describes how we use Point Convolutional Neural Networks (PCNN) proposed in \cite{atzmon2018point} to learn per-point conditional features.

\textbf{Point Cloud Convolutional Operation}
Atzmon et al. \cite{atzmon2018point} propose a method that generalizes convolution operation to points with \textit{extension} and \textit{restriction} operators. The point convolution is done by transforming points to the continuous 3D domain by \textit{extension} $\mathcal{E_{\mathit{P}}}$, applying continuous convolution, and then transforming back to points domain with \textit{restriction} $\mathcal{R}_{\mathit{P}}$, i.e.
\begin{equation} \label{PCNN_operation}
O_{P}=\mathcal{R}_{\mathit{P}}\circ \mathcal{O}_{\mathcal{K}} \circ\mathcal{E_{\mathit{P}}}.
\end{equation}

Raw point cloud input is treated as the sum of a cluster of unit impulse functions, that each impulse is located at 3D Cartesian coordinates of each point. Suppose the point cloud contains $I$ points, it's treated as a set of tuples, each tuple has two attributes: 3D coordinate of the point and the impulse intensity at this location, and denoted as $\left\{ \mathbb{R}^{I\times3},\mathbb{R}^{I\times J}\right\}$. At input layer, $J=1$ indicating the raw point cloud is a sequence (length being $I$) of unit impulses. At hidden layers, $F \in \mathbb{R}^{I\times J}$ represents the union of $J$-dimensional features at each point.

Let $F \in \mathbb{R}^{I\times J}$ be a sparse sampling (at $I$ points) of a 3D field. Extension operation $\mathcal{E_{\mathit{P}}}$ maps it back to a continuous volumetric function, i.e.,
\begin{equation}
\mathcal{E}_{\mathit{P}}:\left\{ \mathbb{R}^{I\times3},\mathbb{R}^{I\times J}\right\} \rightarrow C\left(\mathbb{R}^{3},\mathbb{R}^{J}\right).
\end{equation}
This operation is done by filtering the sequence of impulse with a Gaussian kernel, equivalent to overlay and sum $I$ Gaussian functions centered at the point $P \in \mathbb{R}^{I\times 3}$ and weighted by $j$th function value in $F \in \mathbb{R}^{I\times J}$. This operation is analogous to interpolation as it turns a sequence of discrete values into a continuous function.

The second operator $O$ in operation (\ref{PCNN_operation}) indicates a volumetric convolution:
\begin{equation}
\mathcal{O}_{\mathcal{K}}:C\left(\mathbb{R}^{3},\mathbb{R}^{J}\right)\rightarrow C\left(\mathbb{R}^{3},\mathbb{R}^{K}\right).
\end{equation}
This operation involves $J\times K$ (corresponding to the feature depths of the predecessor and successor layers) filtering kernels. Each kernel is implemented as a set of Gaussian functions whose centers are arranged as a $3\times 3\times 3$ grids. The magnitudes of each Gaussian are learned parameters.

The restriction operation simply samples the function value at points in $P$, turning it back to discrete.
\begin{equation}\label{Restriction_operation}
\mathcal{\mathcal{R}}_{\mathit{P}}:C\left(\mathbb{R}^{3},\mathbb{R}^{K}\right)\rightarrow \left\{ \mathbb{R}^{I\times3},\mathbb{R}^{I\times K}\right\}.
\end{equation}

With the notions above, point convolution operation in (\ref{PCNN_operation}) is abstracted as:
\begin{equation}
\mathcal{O}_{P}:\left\{ \mathbb{R}^{I\times3},\mathbb{R}^{I\times J}\right\}\rightarrow\left\{ \mathbb{R}^{I\times3},\mathbb{R}^{I\times K}\right\},\:\left\Vert P\right\Vert =I,
\end{equation}
which accomplishes a feature mapping ($J$-dimensional to $K$-dimensional) for each point. Similar to the counterparts in image CNNs, Atzmon et al. \cite{atzmon2018point} define pooling and feature map expansion (deconvolution) layers by sampling either a subset (pooling) or a superset (expansion) of $P$ in restriction operation $\mathcal{R}_{\left( \cdot \right)}$.

Since the result of Gaussian convolution is still Gaussian function, $O_{P}$ has a closed-form solution and could be computed efficiently.

\begin{figure}[t!]
	\includegraphics[width=\linewidth]{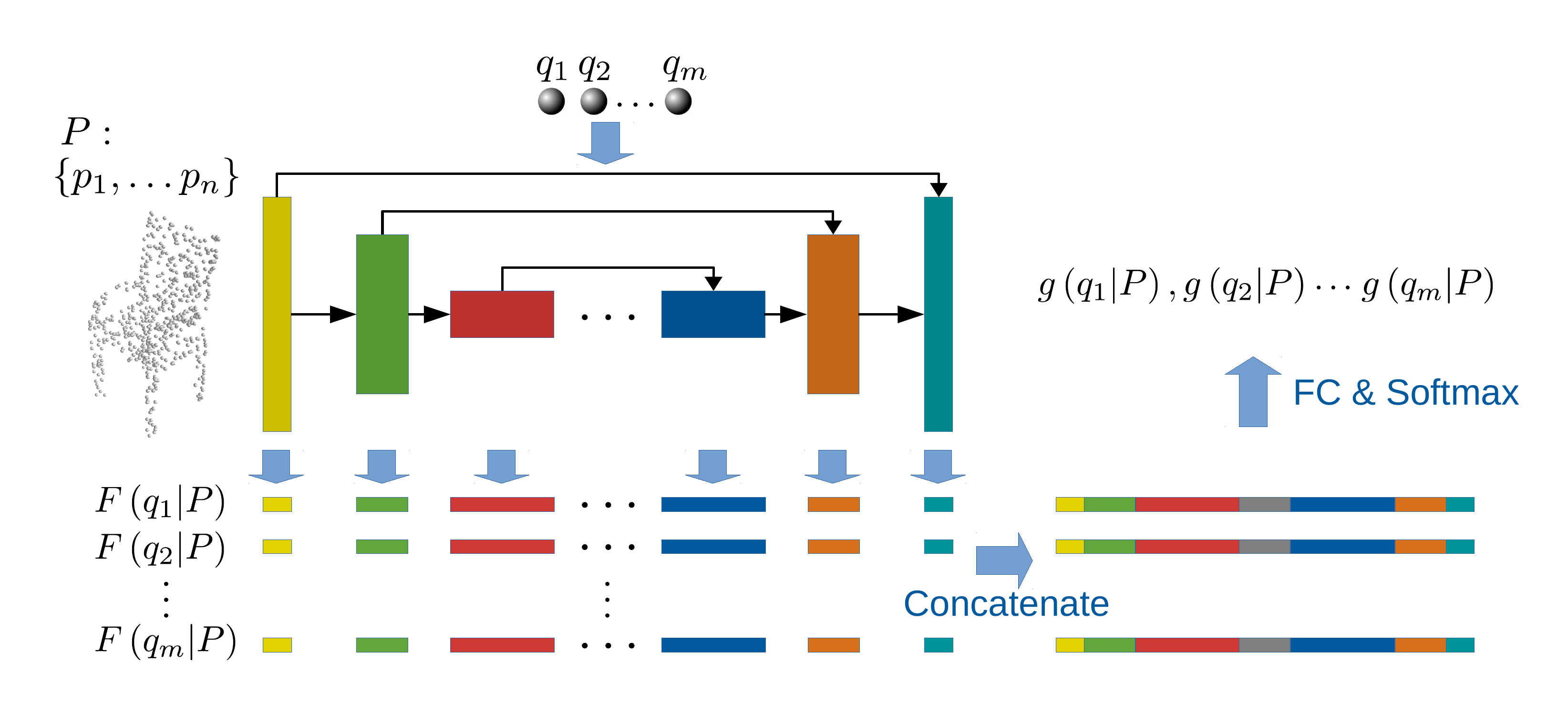}
	\caption{Schematic architecture of our occupancy learning network.}
	\label{fig_Archetecture_of_Occupancy_Learning}
\end{figure}

\textbf{Dual-restriction Point Convolution}
Note we can sample the continuous function at any arbitrary point to apply restriction operation (\ref{Restriction_operation}). If the purpose is to learn and predict a spacial property, like occupancy or SDF, we should take a sample at the query point $q$ where we need its occupancy likelihood. Two distinct restriction operators are used in our point convolution unit. One of them collects features to the point cloud $P$ for feed-forward information flow, just like the original PCNN does. And the other collects to query points $Q$ for gathering shape features to make occupancy inference. This dual-restriction point convolution is denoted as
\begin{equation}
\left\{ O_{P},O_{P\rightarrow Q}\right\},
\end{equation}
where
\begin{equation*}
\begin{array}{c}
O_{P\rightarrow Q}=\mathcal{R}_{\mathit{Q}}\circ \mathcal{O}_{\mathcal{K}}\circ\mathcal{E_{\mathit{P}}},\\
O_{P\rightarrow Q}:\left\{ \mathbb{R}^{I\times K},\mathbb{R}^{I\times K}\right\} \rightarrow\left\{ \mathbb{R}^{L\times3},\mathbb{R}^{L\times K}\right\} ,\;\left\Vert Q\right\Vert =L.
\end{array}
\end{equation*}
$O_{P}$ and $O_{P\rightarrow Q}$ can either sharing the same convolution kernels in $\mathcal{O}_{\mathcal{K}}$ or having distinct ones. In our experiment, we let them share the same set of filtering kernels to avoid extra complexity, as the model provides enough capacity and satisfactory performance.

\textbf{Architecture of PCNN-based Occupancy Learning}
Our occupancy prediction network is constructed with dual-restriction convolution units. Its architecture is shown in Fig. \ref{fig_Archetecture_of_Occupancy_Learning}. Similar to point cloud segmentation, occupancy learning is also a point-level discriminative task. So we use an architecture inspired by U-Net \cite{ronneberger2015u} and adapted from the segmentation network in \cite{atzmon2018point}. It consists of 7 dual-restriction convolution blocks, each connected by feed-froward and skip connections. Skip connections that feed point cloud features to high-level layers may enhance fine details of learned shapes. Those broad blue arrows indicate the side outputs, $O_{P\rightarrow Q}$, of our dual-restriction layers. Query points are feed to each dual-restriction unit, producing seven multi-scale feature vectors for each $q_{i}$. Those features are concatenated to form a single vector for each point, $F\left(q|P \right)$. As mentioned in section \ref{OFL_SUBSEC}, we use a fully-connected layer and Softmax to map $F\left(q|P \right)$ to final output $F\left(q|P \right)$. Loss function used to train this network is defined as (\ref{occu_learning_loss}).

To infer whether a point is inside or outside of a closed surface, one should look at its surroundings, like the distance to the input point clusters ($\mathcal{E_{\mathit{P}}}$) or combinations of shape features ($\mathcal{O}_{\mathcal{K}}$). Then a score ($\mathcal{R}_{\mathit{Q}}$) that's relevant to occupancy is assigned to this query point. Our network has a perpendicular filter-through structure. The information of theinput point cloud is flowing "horizontally", and the query points are feed "vertically". Dual-restriction layers work as an interactor that makes query points entangled with the shape information, getting \emph{conditional} features $F\left(q|P \right)$.

\subsection{Shape Reconstruction from Occupancy Function} \label{sec_SROF}
Occupancy network $g\left(q\mid P \right)$ is an analog representation of the implicit shape of $P$ (cf. Figure \ref{fig_system}). It has to be digitized to be able to rendered or saved. Sometimes an upsampled point cloud is produced as the output of reconstruction (like in point cloud upsampling \cite{yifan2019patch, yu2018pu} or consolidation \cite{yu2018ec, roveri2018pointpronets}). By linear-search over space and pickup points that just cross the decision boundary, we could collect a virtually infinite number of points on the iso-surface. If mesh output is preferred, we can sample a voxel grid over the space and build iso-surface with Marching Cubes \cite{lewiner2003efficient}. 

To compare our occupancy learning methods with other surface reconstruction methods \cite{bernardini1999ball, kazhdan2006poisson, choy20163d, liao2018deep, mescheder2018occupancy} on metrics like Chamfer distance and IoU, we have to produce triangle meshes. So we use the proposed occupancy network as the frontend and construct meshes from predicted implicit functions with Marching Cubes \cite{lewiner2003efficient}. This requires sampling the implicit functions with a grid of points in the bounding box around input point clouds. We can directly sample a $256 \times 256 \times 256$ gird, like in the voxel grid 3D shape representation. That is not feasible as it requires computing network feed-forward pass for 16,777,216 times, without considering batch. Some works \cite{tatarchenko2017octree, hane2017hierarchical} use Octree hierarchical structure representations to alleviate the computation and memory cost in volumetric 3D shape learning. Mescheder et al. \cite{mescheder2018occupancy} apply this method sampling implicit function frontend to accelerate surface reconstruction. In the experiment, we use the same way to build three levels of grid $64^{3}\rightarrow128^{3}\rightarrow256^{3}$ and then construct iso-surface with Marching Cubes.

\begin{figure*}[h!]
	\centering
	\includegraphics[width=\linewidth]{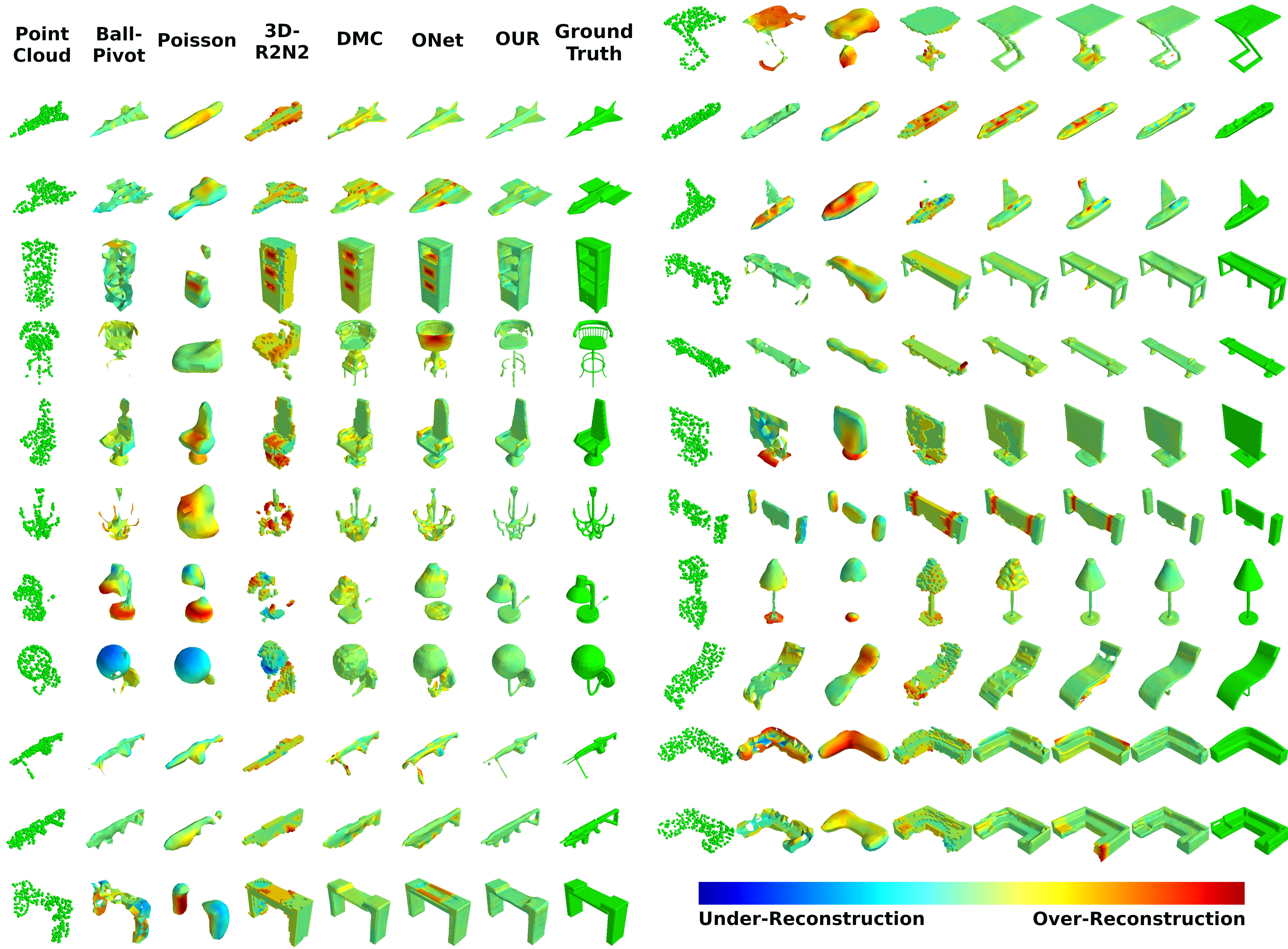}
	\caption{Comparison of our method with Ball-Pivot, Poisson method, 3D-R2N2, DMC, and ONet. For making it easier to compare with the ground truth, we use jet colormap to visualize the reconstruction errors. The surfaces reconstructed out of the ground truth is colored in red, and inside of the ground truth colored in blue. We scale the error across each example to maximize the color range, but keep it linear and keep zero-error colored in light-green constantly.}
	\label{fig_compare}
\end{figure*}

\section{Implementation details and Experiments} \label{Sec_experiments}

\subsection{3D Mesh Reconstruction of ShapeNet}\label{sec_3DMRS}
In this experiment, we evaluate our occupancy learning approach with the task of reconstructing shapes from unoriented point clouds. We take a subset of ShapeNet \cite{chang2015shapenet} (as the same of 3D-R2N2 \cite{choy20163d} and Occupancy-Network \cite{mescheder2018occupancy}) and preprocess each mesh to make it watertight with the code provided by \cite{stutz2018learning}. We refer to it as ShapeNet-13 hereafter, as it consists of 13 classes of shapes. Our training/validation/testing split is the same as \cite{mescheder2018occupancy}. All of the meshes are rescale and shifted into a unit cube centered at the origin of coordinate system. 2048 points are randomly sampled from each mesh surface for data augmentation. Each input point cloud $P$ contains 300 points randomly taken from these 2048 points, and Gaussian noise is applied to each point. For generating query points (cf. Sec. \ref{OFL_SUBSEC}), we add random offsets to on-surface points to make them off-surface. In practice, we draw 6144 points with $\delta=0.02$ normal distribution to capture details of shape and draw 2048 points with $\delta=0.1$ that reflecting the overall shape contour.

The architecture of our network is inspired by the segmentation model in \cite{atzmon2018point}. As shown in Figure \ref{fig_Archetecture_of_Occupancy_Learning}, the input point cloud is feed through a cascade of shrinking and enlarging streams. At each convolution block of the shrinking stream, we decrease the size of point cloud (with point max-pooling \cite{atzmon2018point}) and double the depth of features. The opposite is done in the enlarging stream, with point deconvolution \cite{atzmon2018point}. The skipping feed-forward features from the shrinking stream are concatenated to the outputs of corresponding deconvolution layers. These size hyper-parameters are summarized as below:

{\tiny\
\begin{equation*}
\overrightarrow{\left(300,64,256\right)\rightarrow\overrightarrow{\left(128,128\right)\rightarrow\overrightarrow{\left(256,16\right)\Rightarrow\left(256,128\right)}\rightarrow\left(128,256\right)}\rightarrow\left(256,300\right)},
\end{equation*}}
where in each block, excepting the first one, is interpreted as (\#feature, \#output points), and the first one as (\#input points, \#feature, \#output points)

After training, our network is used as a frontend, and we use Octree hierarchical sampling and Marching Cube to produce triangle meshes from each point cloud. We compare our approach against other reconstruction or shape generation methods, including Ball-Pivoting \cite{bernardini1999ball}, Poisson method \cite{kazhdan2006poisson}, 3D-R2N2 \cite{choy20163d}, DMC \cite{liao2018deep}, and ONet \cite{mescheder2018occupancy}.

\begin{table}
	\caption{Shape Reconstruction Results on ShapeNet-13 Dataset, 300 Points, Noise SD=0.05}
	\label{table_1}
	\centering
	\begin{tabular}{cccc}
		\toprule
		{} & {IoU} & {Chamfer-L1} & {Normal} {Consistency}\\
		\midrule
		{Ball-Pivoting} & {-} & {0.415} & {0.720}\\
		{Poisson} & {0.349} & {0.534} & {0.717}\\
		{3D-R2N2} & {0.565} & {0.169} & {0.719}\\
		{DMC} & {0.674} & {0.117} & {0.848}\\
		{ONet} & {0.778} & {0.079} & {0.895}\\
		{Our} & \textbf{0.817} & \textbf{0.060} & \textbf{0.905}\\
		\bottomrule
	\end{tabular}
\end{table}

\begin{table*}[h!]
	\caption{Per-class Statistic of Reconstruction Results on ShapeNet-16 Dataset, 300 Points, Noise SD=0.05}
	\label{table_2}
	\centering
	\begin{tabular}{cc|{c}p{0.7cm}{c}p{0.7cm}{c}p{0.7cm}{c}p{0.7cm}{c}p{0.7cm}{c}p{0.7cm}{c}p{0.7cm}{c}p{0.7cm}{c}p{0.7cm}{c}p{0.7cm}{c}p{0.7cm}{c}p{0.7cm}{c}p{0.7cm}{c}p{0.7cm}{c}}
		\toprule 
		& \multicolumn{1}{c}{category} & airplane & bench & cabinet & car & chair & display & lamp & speaker & rifle & sofa & table & phone & vessel & \textbf{mean} \tabularnewline
		\hline 
		\multirow{3}{*}{\rotatebox[origin=c]{90}{class} \rotatebox[origin=c]{90}{size}} & train & 2832 & \hfil 1101 & 4746 & \hfil 1624 & 1661 & \hfil 5958 & 1359 & \hfil 1272 & 5248 & \hfil 767 & 1134 & \hfil 2222 & 737 & \hfil - \tabularnewline
		& val & 404 & \hfil 157 & 677 & \hfil 231 & 237 & \hfil 850 & 193 & \hfil 181 & 749 & \hfil 109 & 161 & \hfil 317 & 105 & \hfil- \tabularnewline
		& test & 809 & \hfil 314 & 1355 & \hfil 463 & 474 & \hfil 1701 & 387 & \hfil 363 & 1499 & \hfil 219 & 323 & \hfil 634 & 210 & \hfil- \tabularnewline
		\hline 
		\multirow{6}{*}{\rotatebox[origin=c]{90}{IoU}} & Ball-Pivoting & \hfil- & \hfil- & \hfil- & \hfil- & \hfil- & \hfil- & \hfil- & \hfil- & \hfil- & \hfil- & \hfil- & \hfil- & \hfil - & \hfil - \tabularnewline
		& Poisson & 0.388 & 0.241 & 0.450 & 0.543 & 0.211 & 0.328 & 0.268 & 0.415 & 0.485 & 0.426 & 0.163 & 0.458 & 0.536 & 0.378 \tabularnewline
		& 3D-R2N2 & 0.459 & 0.425 & 0.717 & 0.673 & 0.505 & 0.585 & 0.328 & 0.701 & 0.413 & 0.698 & 0.468 & 0.665 & 0.542 & 0.552 \tabularnewline
		& DMC & 0.546 & 0.523 & 0.798 & 0.736 & 0.636 & 0.740 & 0.480 & 0.791 & 0.563 & 0.798 & 0.611 & 0.840 & 0.608 & 0.667 \tabularnewline
		& ONet & 0.757 & \textbf{0.715} & \textbf{0.861} & \textbf{0.831} & 0.731 & 0.813 & 0.564 & \textbf{0.823} & 0.690 & 0.867 & 0.755 & 0.910 & 0.745 & 0.774\tabularnewline
		& Our & \textbf{0.800} & 0.694 & 0.855 & 0.816 & \textbf{0.765} & \textbf{0.852} & \textbf{0.687} & 0.814 & \textbf{0.782} & \textbf{0.885} & \textbf{0.774} & \textbf{0.931} & \textbf{0.800} & \textbf{0.804} \tabularnewline
		\hline 
		\multirow{6}{*}{\rotatebox[origin=c]{90}{Chamfer-L1} \rotatebox[origin=c]{90}{(less is better)}} & Ball-Pivoting & 0.245 & 0.241 & 0.544 & 0.251 & 0.575 & 0.446 & 0.467 & 0.711 & 0.075 & 0.401 & 0.633 & 0.209 & 0.164 & 0.382 \tabularnewline
		& Poisson & 0.423 & 0.400 & 0.579 & 0.287 & 0.817 & 0.571 & 0.590 & 0.744 & 0.193 & 0.454 & 0.756 & 0.318 & 0.240 & 0.490 \tabularnewline
		& 3D-R2N2 & 0.153 & 0.157 & 0.171 & 0.201 & 0.202 & 0.180 & 0.283 & 0.203 & 0.144 & 0.161 & 0.187 & 0.145 & 0.181 & 0.182 \tabularnewline
		& DMC & 0.111 & 0.102 & 0.114 & 0.165 & 0.121 & 0.099 & 0.167 & 0.150 & 0.083 & 0.102 & 0.103 & 0.064 & 0.147 & 0.118 \tabularnewline
		& ONet & 0.056 & 0.059 & \textbf{0.073} & 0.098 & 0.089 & 0.076 & 0.138 & 0.116 & 0.060 & 0.069 & 0.071 & 0.041 & 0.085 & 0.079 \tabularnewline
		& Our & \textbf{0.045} & \textbf{0.057} & 0.076 & \textbf{0.076} & \textbf{0.076} & \textbf{0.062} & \textbf{0.078} & \textbf{0.107} & \textbf{0.039} & \textbf{0.058} & \textbf{0.064} & \textbf{0.032} & \textbf{0.058} & \textbf{0.064} \tabularnewline
		\hline 
		\multirow{6}{*}{\rotatebox[origin=c]{90}{Normal} \rotatebox[origin=c]{90}{Consistency}} & Ball-Pivoting & 0.753 & 0.704 & 0.719 & 0.703 & 0.669 & 0.739 & 0.717 & 0.726 & 0.811 & 0.702 & 0.716 & 0.799 & 0.785 & 0.734 \tabularnewline
		& Poisson & 0.702 & 0.661 & 0.768 & 0.763 & 0.656 & 0.741 & 0.690 & 0.766 & 0.786 & 0.743 & 0.677 & 0.802 & 0.763 & 0.732 \tabularnewline
		& 3D-R2N2 & 0.651 & 0.683 & 0.784 & 0.714 & 0.667 & 0.749 & 0.583 & 0.746 & 0.676 & 0.751 & 0.728 & 0.841 & 0.626 & 0.708 \tabularnewline
		& DMC & 0.804 & 0.802 & 0.886 & 0.825 & 0.837 & 0.900 & 0.769 & 0.880 & 0.771 & 0.885 & 0.870 & 0.947 & 0.797 & 0.844 \tabularnewline
		& ONet & 0.897 & \textbf{0.878} & \textbf{0.915} & \textbf{0.874} & \textbf{0.889} & 0.925 & 0.812 & \textbf{0.897} & 0.863 & \textbf{0.927} & \textbf{0.916} & 0.970 & 0.859 & 0.894 \tabularnewline
		& Our & \textbf{0.910} & 0.858 & 0.904 & 0.848 & 0.885 & \textbf{0.928} & \textbf{0.859} & 0.881 & \textbf{0.902} & 0.925 & 0.909 & \textbf{0.974} & \textbf{0.880} & \textbf{0.897} \tabularnewline
		\bottomrule 
	\end{tabular}
\end{table*}

The reconstruction quality is evaluated in Intersection-over-Union (IoU), average L1 Chamfer distance (Chamfer-L1), and Normal Consistency score, defined as below:
\begin{enumerate}
	\item Intersection-over-Union: Quotient of output mesh and ground truth mesh's intersection's volume and their union's, i.e., $\mathrm{vol}\left(A\cap B\right)/\mathrm{vol}\left(A\cup B\right)$.
	\item L1 Chamfer distance: Sum of an \emph{accuracy measurement}, mean distance from points on the output mesh to ground-truth, and a \emph{completeness measurement}, the mean distance from ground-truth mesh to output. Formally, $\sum_{x\in A}\mathrm{min_{y\in B}}\left|x-y\right|/N_{x}+\sum_{y\in B}\mathrm{min_{x\in A}}\left|x-y\right|/N_{y}$.
	\item Normal consistency score: Normal consistency score consists of two terms, \emph{normal accuracy} and \emph{normal completeness}. They are the mean absolute dot product of the surface normals on output mesh and the surface normals at the corresponding nearest points on the ground-truth mesh, and the opposite. Formally, $\sum_{x\in A,y=\mathrm{argmin}_{y\in B}\left\Vert x-y\right\Vert }N(x)\cdot N(y)+\sum_{y\in B,x=\mathrm{argmin}_{x\in A}\left\Vert x-y\right\Vert }N(x)\cdot N(y)$.
\end{enumerate}

We report the reconstruction performance in Table \ref{table_1}. We find that our method outperforms all other methods on IoU and Chamfer-L1 metrics by a large gap but only is marginally better than ONet/DMC on normal consistency score. For visual inspection and evaluation, we plot the output meshes of different methods and visualizing surface error in Figure \ref{fig_compare}. The error is color-coded, where the reconstructed surface out of ground-truth is colored in red, and underneath is colored in blue. The dataset contains 13 different classes of objects. Those class names and per-class statistics of the performance are shown in Table \ref{table_2}.

\subsection{Difference between our method and ONet}

\begin{figure}[h]
	\centering
	\subfloat[ONet: good; Our: good\label{fig_compare_detail_a}]{
		\includegraphics[width=0.45\linewidth]{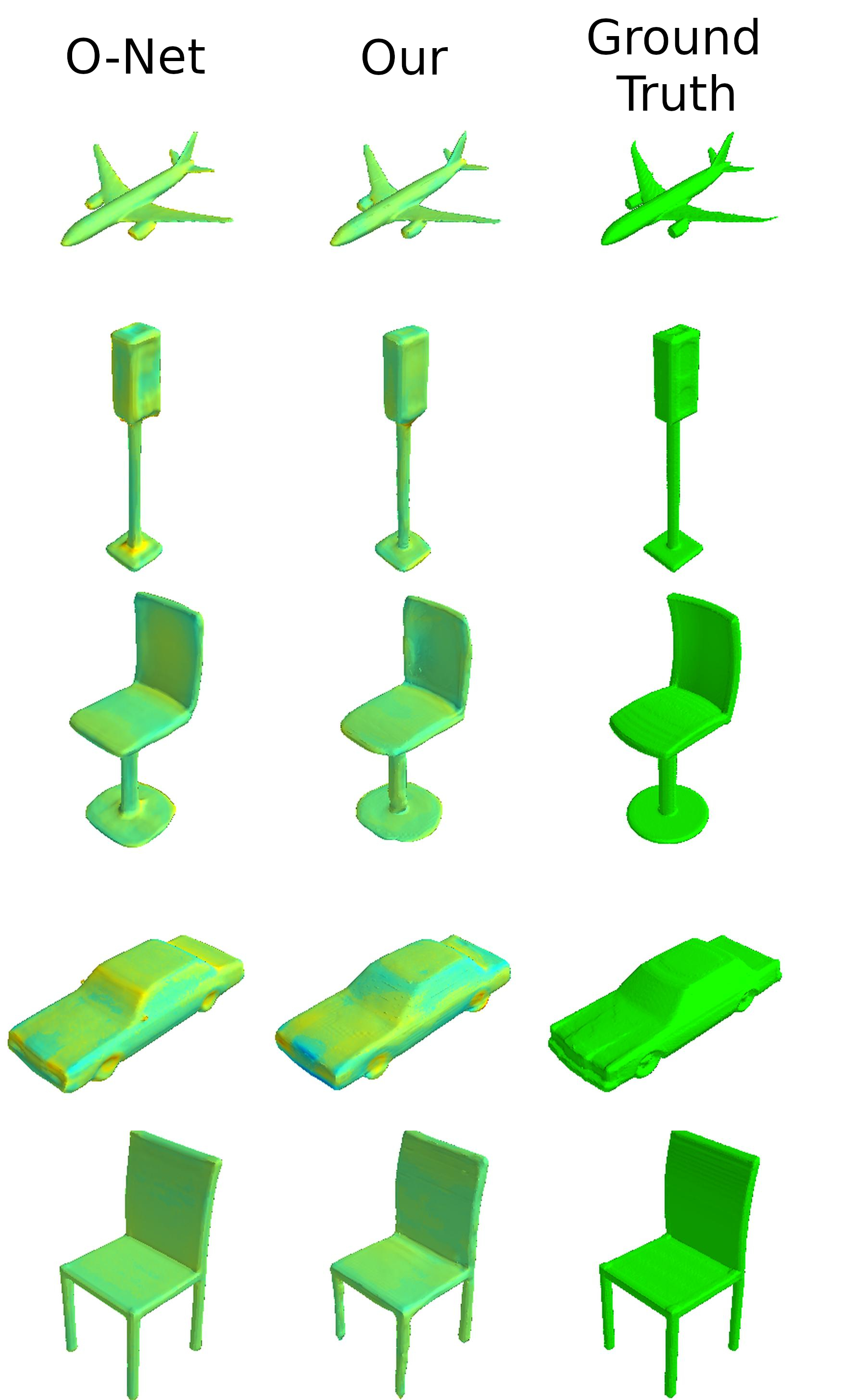}}
	\subfloat[ONet: good; Our: bad\label{fig_compare_detail_b}]{
		\includegraphics[width=0.45\linewidth]{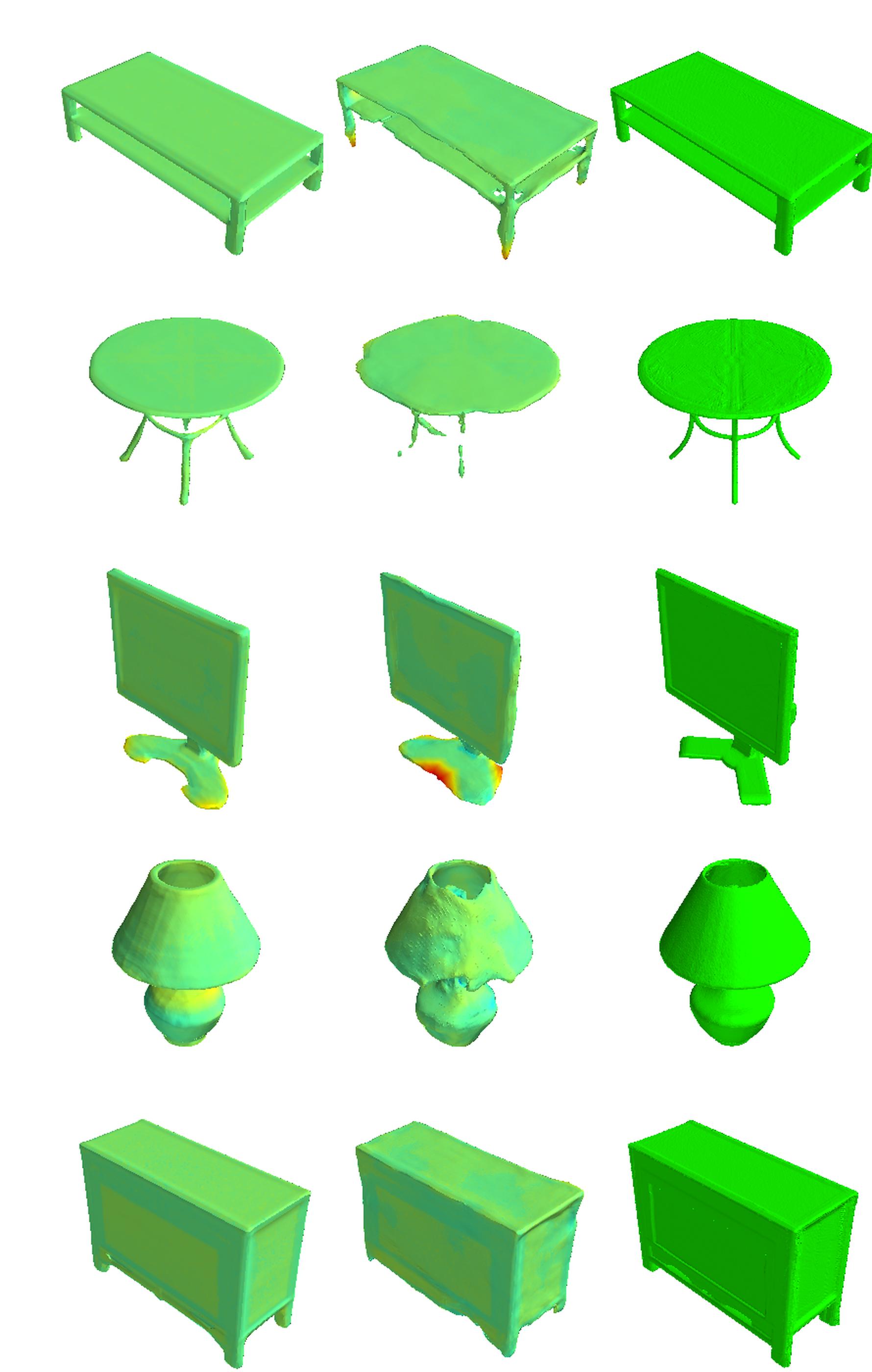}}\\
	\subfloat[ONet: bad; Our: good\label{fig_compare_detail_c}]{
		\includegraphics[width=0.45\linewidth]{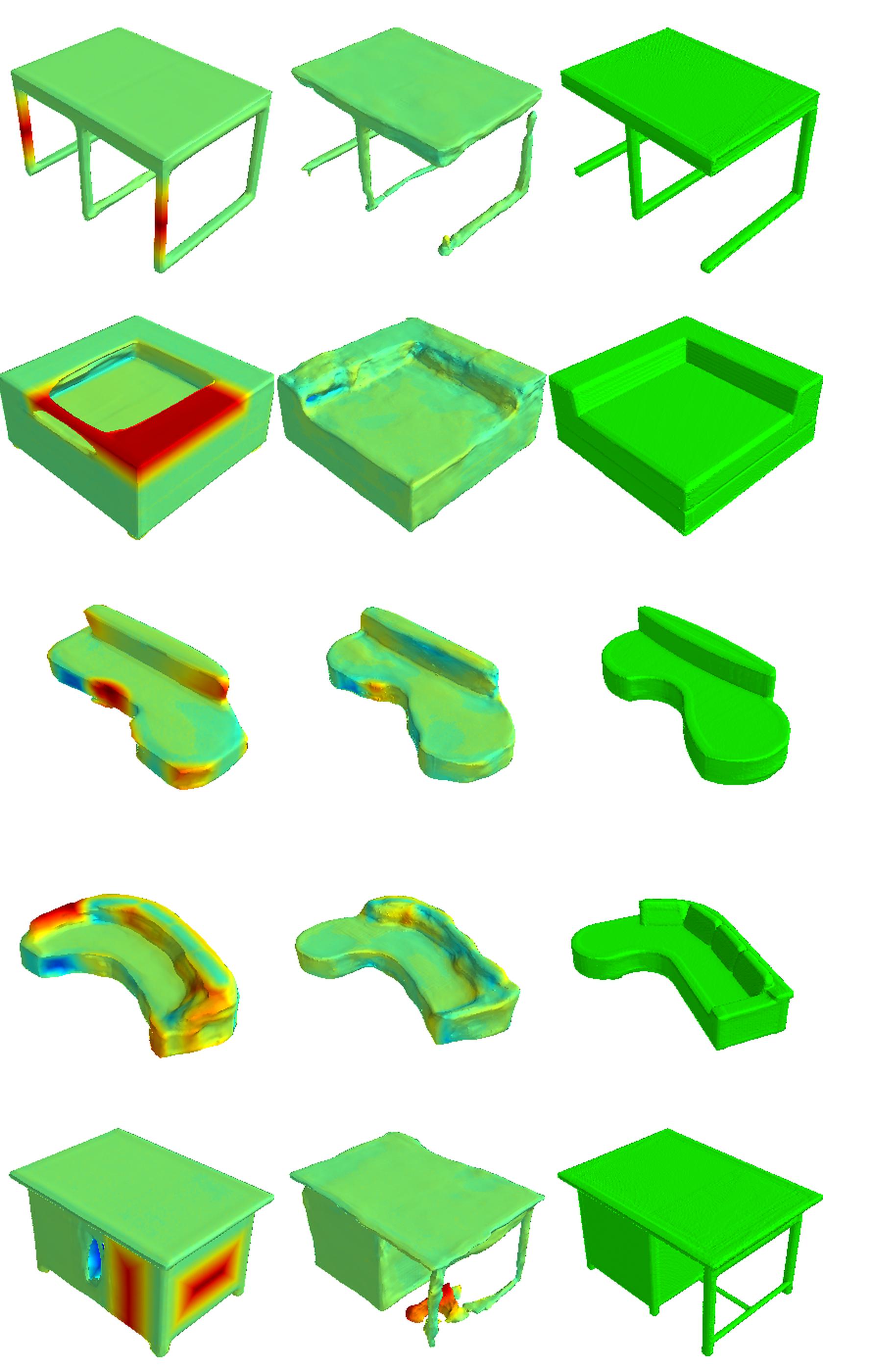}}
	\subfloat[ONet: bad; Our: bad\label{fig_compare_detail_d}]{
		\includegraphics[width=0.45\linewidth]{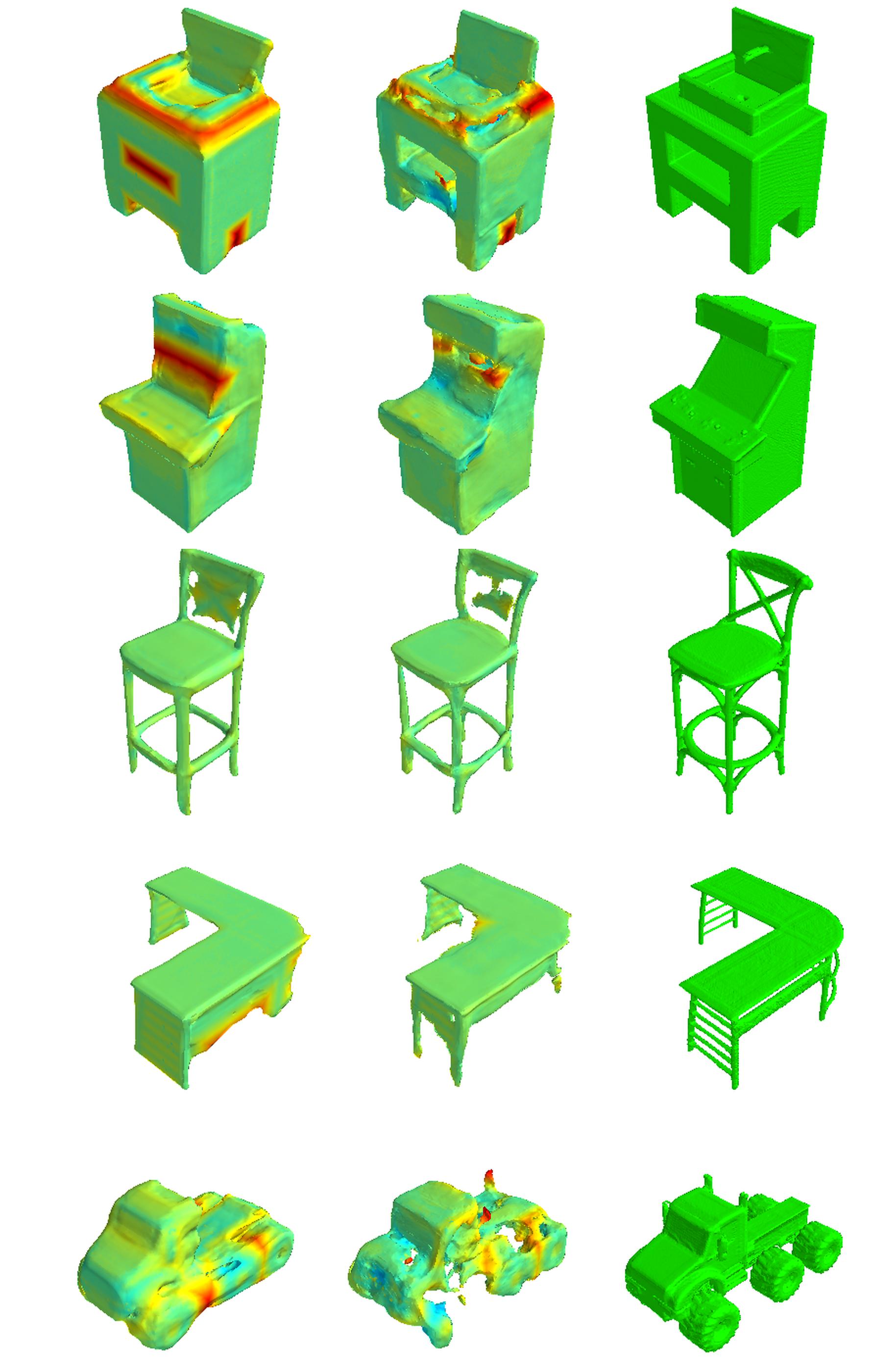}}
	\caption{Comparison of our method with ONet. We collect four groups of test shapes according to if it's reconstructed well by ONet and Our method.}
	\label{fig_compare_detail}
\end{figure}

Among the methods compared, ONet \cite{mescheder2018occupancy} is the one closest to our method in terms of quantitative metrics. ONet produces some good-looking reconstruction in Fig. \ref{fig_compare} and seems to have advantages for learning from some particular classes of shapes (Table \ref{table_2}). We are therefore interested in its behavior and the differences from our PCNN-based occupancy learning. Browsing through output meshes of ONet and our, we select some representative shapes and group them as a) both work well; b) ONet works better; c) our method performs better; d) doth work not well. See Fig. \ref{fig_compare_detail} for some examples in these four groups. There are at least three noticeable characters worth mention:
\begin{enumerate}
	\item First, ONet is good at learning and recovering the classes that have less variation in shape geometry. Shape classes like airplanes, tables, and cars are examples of "less variant" categories, as you can easily imagine a stereotypical of them (cf. Fig. \ref{fig_compare_detail_a}, \ref{fig_compare_detail_b} and Table \ref{table_2}). ONet uses an encoder-decoder network using a PointNet \cite{qi2017pointnet} encoder to extract a latent code from the input point cloud. As a result, the categorical and semantic features are captured relatively well, but this brings byproducts and problems. ONet tends to remember stereotypical shapes and inclines to fit every input shape to those remembered corpus, resulting in imaginary fake shapes (Fig. \ref{fig_compare_detail_c}) and fails when working with unseen structures.
	
	\item Second, ONet produces cleaner and more regular mesh than our methods, especially for the shapes that could be approximated as Assembly of primary elements like planes, cubes, circles, and discs (cf. Fig. \ref{fig_compare_detail_b}). While our outputs, even for the shapes reconstructed better the ONet, contain more noise and non-smooth surfaces. This observation explains our method's relative weakness on the snormal consistency metric in Table \ref{table_1}. This phenomenon is a result that we use FBF-based PCNN (local and grained) to model the occupancy function, while ONet uses a conditional batch normalization \cite{dumoulin2016adversarially} network as the decoder. However, we believe scarifying regularity and smoothness is worthy as our method can better reconstruct the shape in-the-wild. Show in Fig. \ref{fig_compare_detail_c}, they are either deviate much from its category (like the desk with horizontal feet and squared sofa) or consist of long curved surfaces and global asymmetries. Note that high frequency \footnote{The noise is also roughly constant frequency, as the wave length is proportional to the RBF kernel size and interval between RBF centres} jitter on the reconstructed surface is easy to remove with mesh spectral filtering.
	
	\item Lastly, some challenging shapes neither ONet nor our method can recover them satisfactorily from 300 input points with noise (Fig. \ref{fig_compare_detail_d}). The reasons are two-fold. First, these objects are composed of multiply complex parts so that 300 oriented points do not provide enough information about their shape. Meanwhile, the training dataset does not contain enough similar objects for these shape corpus, so that our network may not recover or "hallucinate" missing information.
\end{enumerate}

\begin{figure}[h]
	\centering
	\includegraphics[width=0.89\linewidth]{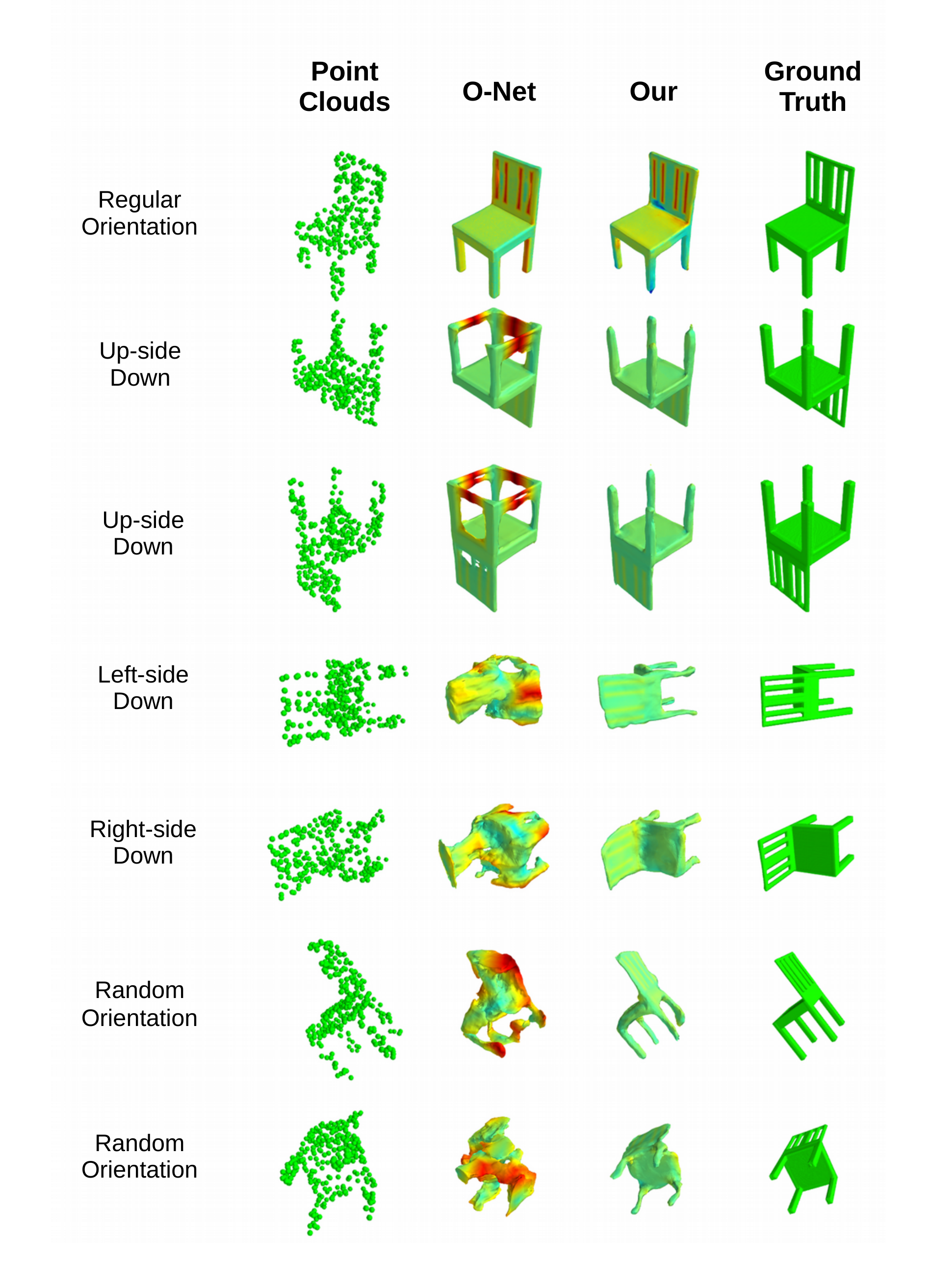}
	\caption{The effects of applying geometric rotations to input point clouds. This experiment shows that a simple small rotation can crush ONet entirely, while our model is relatively invariant to rotation.}
	\label{fig_up_down}
\end{figure}

Traditional surface reconstruction methods like MLS \cite{fleishman2005robust} and Poisson method \cite{kazhdan2006poisson} are invariant to geometric translation and rotation. Unfortunately, most recent point cloud deep learning frameworks, including those learning-based ones in Table \ref{table_1} do not have such property. We intuitively feel ONet relies more on a canonical pose of the input point cloud, as recognizing and fitting stereotypical shape is pose-related. Fig. \ref{fig_up_down} shows an experiment that we deliberately rotate a chair's point cloud into different poses to test ONet and our method. While both of these methods produce a success reconstruction in regular orientation, ONet fails to react robustly to the rotations. In sum, our occupancy learning method is more faithful to the input shape information. The shape detail is recovered more from local shape evidence than what ONet and shape generation methods are doing: remembering and fitting stereotypical shapes of the categories.

\subsection{Scaling and Generalization}

\begin{figure}[t]
	\centering
	\includegraphics[width=\linewidth]{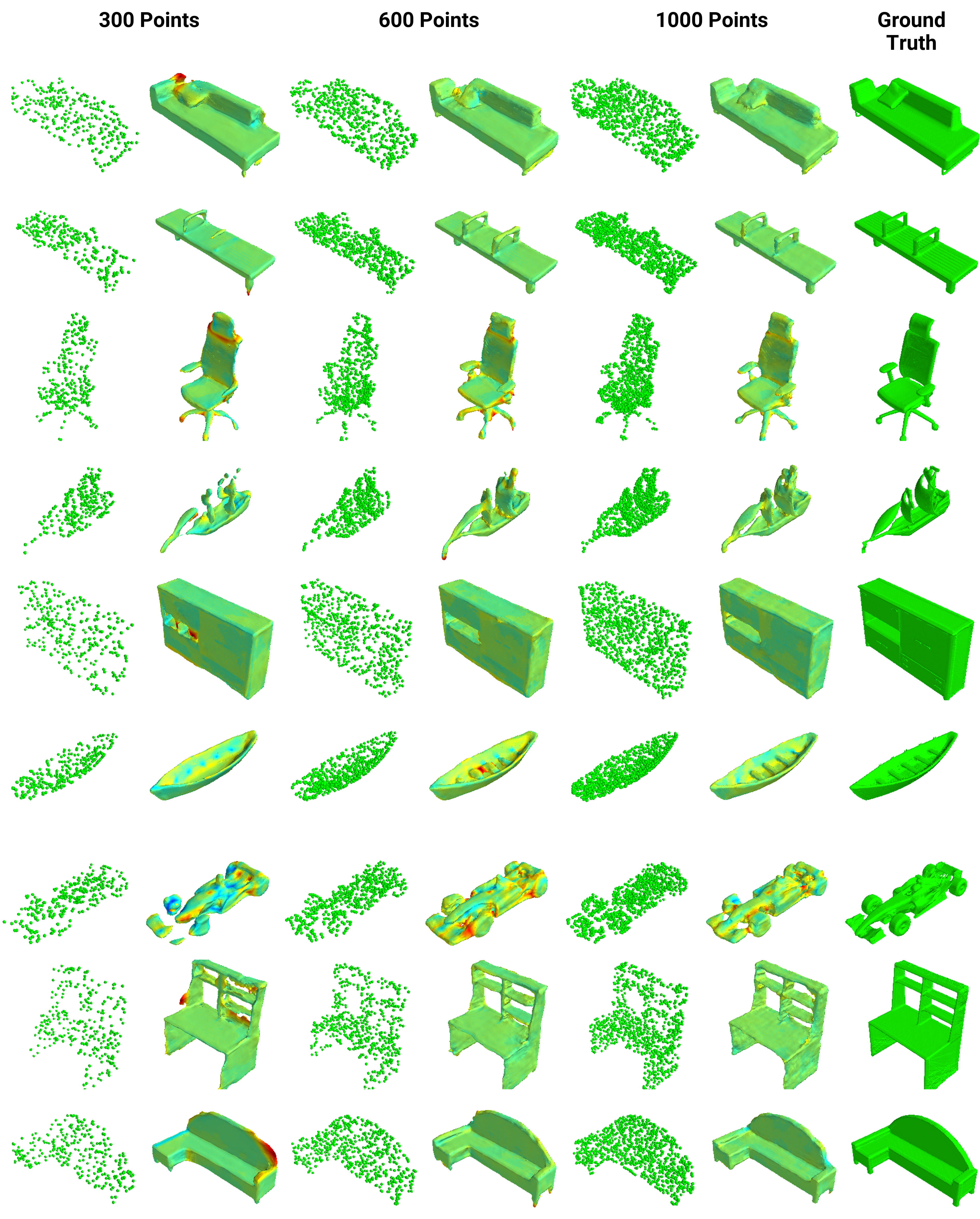}
	\caption{This figure shows the shapes generated from different densities of input point clouds, with correspondingly trained networks}
	\label{fig_N_points}
\end{figure}

Since our method can reliably reconstruct sufficient fine details with local geometric evidence, we expect it to extend to reconstructing denser sample shapes and gaining accuracy from richer information. Meanwhile, the proposed sampling layer for PCNN and the network architecture are easy to scale-up. We conduct several experiments the have varied number/density of input points and different level of noise. The results are reported in Table \ref{table_3}. We show some 3d shapes reconstructed by networks configured and trained with 300/600/1000 input points In Fig. \ref{fig_N_points}. From the metrics and visual results, we can see that reconstruction accuracy and quality increase regularly with denser input and lower noisy intensity. In these experiments, the network architecture or training parameters are simply copied from the initial experiment setups, except for the size of Gaussian kernels used in each PCNN layers, which are set as reciprocal of the number of points of a particular layer.

\begin{table}[h]
	\caption{Performance of our method on ShapeNet-13 Dataset, with different density and noise}
	\label{table_3}
	\centering
	\begin{tabular}{cc|ccc}
		\toprule
		\multicolumn{2}{c}{Density \& Noise} & {IoU} & {Ch.L1} & {Normal}\\ 
		\midrule
		{\scriptsize{150}} & \scriptsize{0.05} & 0.741 & 0.099 & 0.860 \\
		\hline
		\multirow{2}{*}{\scriptsize{300}} & \scriptsize{0.05} & 0.817 & 0.060 & 0.905 \\
		{} & \scriptsize{0.025} & 0.800 & 0.071 & 0.890 \\
		\hline
		\multirow{2}{*}{\scriptsize{600}} & \scriptsize{0.05} & 0.823 & 0.060 & 0.905 \\
		{} & \scriptsize{0.025} & 0.842 & 0.052 & 0.913 \\
		\hline
		{\scriptsize{1000}} & \scriptsize{0.05} & 0.870 & 0.045 & 0.932 \\
		\bottomrule
	\end{tabular}
\end{table}

\begin{figure}[t!]
	\centering
	\includegraphics[width=\linewidth]{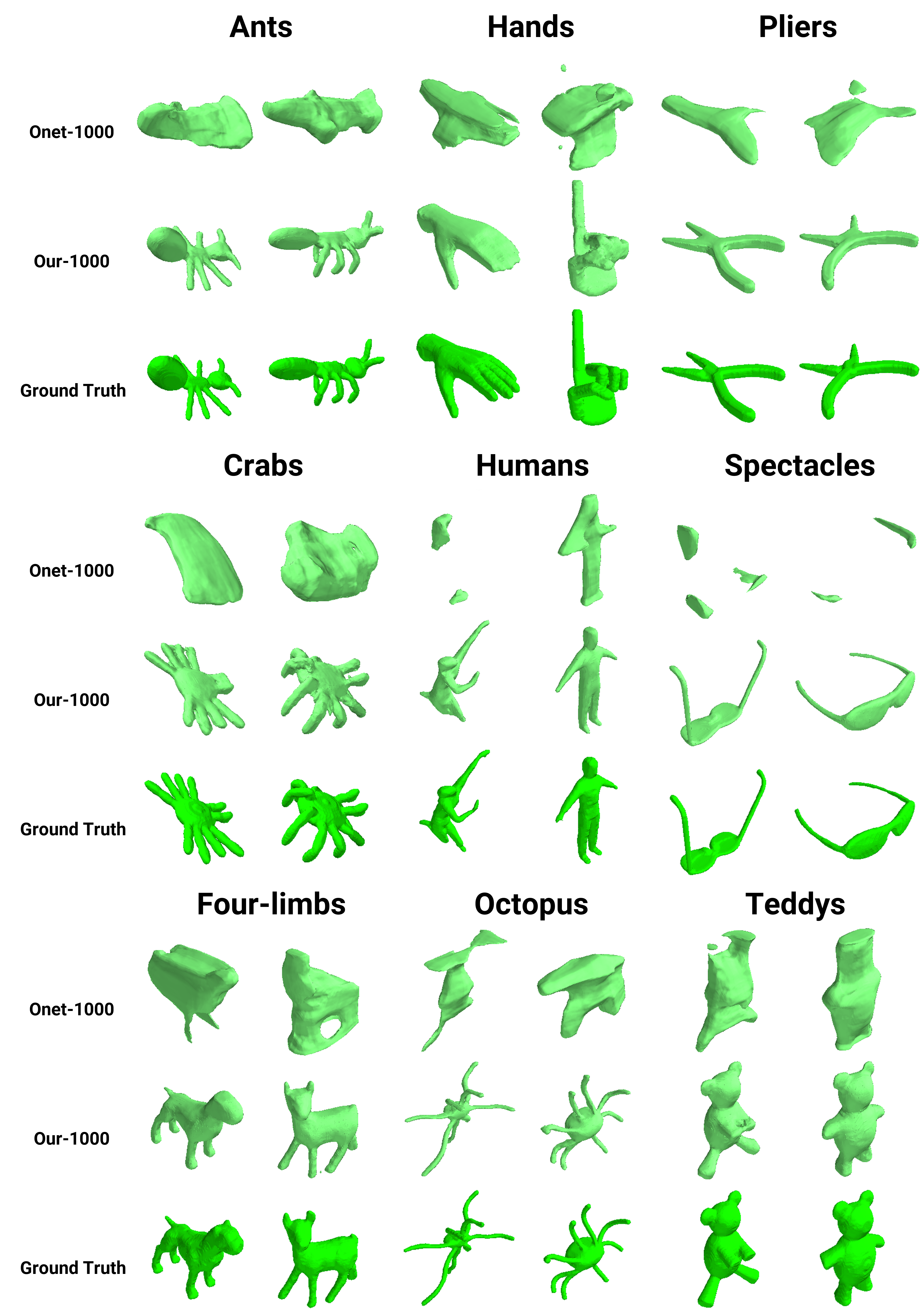}
	\caption{Comparison of our method with ONet on unseen classes in McGill dataset \cite{siddiqi2008retrieving}.}
	\label{fig_McGill}
\end{figure}

To test our method's power of generalization to unseen object categories, we apply the 1000-point model trained on ShapeNet-13 to the McGill mesh dataset \cite{siddiqi2008retrieving}, without any fine-tuning. Point cloud sampling is done in the same way mentioned in \ref{sec_3DMRS}. A 1000-point ONet is trained accordingly to make a comparison. The reconstruction results shown in Fig. \ref{fig_McGill} demonstrate that our method has remarkable potential for up-scaling and domain generalization.

\section{Conclusion and Future Works}
In this paper, we propose a novel occupancy learning method and use it to improve surface reconstruction. Because our occupancy prediction model is built with point convolutional components, it behaves differently and performs better than the previous implicit function learning methods on ShapeNet \cite{chang2015shapenet} reconstruction.

We are optimistic about generalizing this method to learn other continuous property of 3D space other than occupancy for future work, for example, the deformation field in nonrigid 3D reconstruction/registration/part correspondence. Our method could also be used to learn the elastic property of materiel and tissue, helping non-linear shape deformation/animation. Meanwhile, the implicit function's domain seems a candidate for fusing information from multi-modality or different types of data: images, point clouds, voxel grids, and even graphs.

\bibliographystyle{IEEEtran}
\bibliography{IEEEabrv,myReferences}

\begin{thebibliography}{10}
\providecommand{\url}[1]{#1}
\csname url@samestyle\endcsname
\providecommand{\newblock}{\relax}
\providecommand{\bibinfo}[2]{#2}
\providecommand{\BIBentrySTDinterwordspacing}{\spaceskip=0pt\relax}
\providecommand{\BIBentryALTinterwordstretchfactor}{4}
\providecommand{\BIBentryALTinterwordspacing}{\spaceskip=\fontdimen2\font plus
\BIBentryALTinterwordstretchfactor\fontdimen3\font minus
  \fontdimen4\font\relax}
\providecommand{\BIBforeignlanguage}[2]{{%
\expandafter\ifx\csname l@#1\endcsname\relax
\typeout{** WARNING: IEEEtran.bst: No hyphenation pattern has been}%
\typeout{** loaded for the language `#1'. Using the pattern for}%
\typeout{** the default language instead.}%
\else
\language=\csname l@#1\endcsname
\fi
#2}}
\providecommand{\BIBdecl}{\relax}
\BIBdecl

\bibitem{chang2015shapenet}
A.~X. Chang, T.~Funkhouser, L.~Guibas, P.~Hanrahan, Q.~Huang, Z.~Li,
  S.~Savarese, M.~Savva, S.~Song, H.~Su \emph{et~al.}, ``Shapenet: An
  information-rich 3d model repository,'' \emph{arXiv preprint
  arXiv:1512.03012}, 2015.

\bibitem{siddiqi2008retrieving}
K.~Siddiqi, J.~Zhang, D.~Macrini, A.~Shokoufandeh, S.~Bouix, and S.~Dickinson,
  ``Retrieving articulated 3-d models using medial surfaces,'' \emph{Machine
  vision and applications}, vol.~19, no.~4, pp. 261--275, 2008.

\bibitem{chen2019learning}
Z.~Chen and H.~Zhang, ``Learning implicit fields for generative shape
  modeling,'' in \emph{Proceedings of the IEEE Conference on Computer Vision
  and Pattern Recognition}, 2019, pp. 5939--5948.

\bibitem{mescheder2018occupancy}
L.~Mescheder, M.~Oechsle, M.~Niemeyer, S.~Nowozin, and A.~Geiger, ``Occupancy
  networks: Learning 3d reconstruction in function space,'' \emph{arXiv
  preprint arXiv:1812.03828}, 2018.

\bibitem{atzmon2018point}
M.~Atzmon, H.~Maron, and Y.~Lipman, ``Point convolutional neural networks by
  extension operators,'' \emph{arXiv preprint arXiv:1803.10091}, 2018.

\bibitem{lewiner2003efficient}
T.~Lewiner, H.~Lopes, A.~W. Vieira, and G.~Tavares, ``Efficient implementation
  of marching cubes' cases with topological guarantees,'' \emph{Journal of
  graphics tools}, vol.~8, no.~2, pp. 1--15, 2003.

\bibitem{choy20163d}
C.~B. Choy, D.~Xu, J.~Gwak, K.~Chen, and S.~Savarese, ``3d-r2n2: A unified
  approach for single and multi-view 3d object reconstruction,'' in
  \emph{European conference on computer vision}.\hskip 1em plus 0.5em minus
  0.4em\relax Springer, 2016, pp. 628--644.

\bibitem{liao2018deep}
Y.~Liao, S.~Donne, and A.~Geiger, ``Deep marching cubes: Learning explicit
  surface representations,'' in \emph{Proceedings of the IEEE Conference on
  Computer Vision and Pattern Recognition}, 2018, pp. 2916--2925.

\bibitem{rekanos2008shape}
I.~T. Rekanos, ``Shape reconstruction of a perfectly conducting scatterer using
  differential evolution and particle swarm optimization,'' \emph{IEEE
  Transactions on Geoscience and Remote Sensing}, vol.~46, no.~7, pp.
  1967--1974, 2008.

\bibitem{gao2019deepspline}
J.~Gao, C.~Tang, V.~Ganapathi-Subramanian, J.~Huang, H.~Su, and L.~J. Guibas,
  ``Deepspline: Data-driven reconstruction of parametric curves and surfaces,''
  \emph{arXiv preprint arXiv:1901.03781}, 2019.

\bibitem{groueix2018atlasnet}
T.~Groueix, M.~Fisher, V.~G. Kim, B.~C. Russell, and M.~Aubry, ``Atlasnet: A
  papier-m$\backslash$\^{} ach$\backslash$'e approach to learning 3d surface
  generation,'' \emph{arXiv preprint arXiv:1802.05384}, 2018.

\bibitem{amenta2002simple}
N.~Amenta, S.~Choi, T.~K. Dey, and N.~Leekha, ``A simple algorithm for
  homeomorphic surface reconstruction,'' \emph{International Journal of
  Computational Geometry \& Applications}, vol.~12, no. 01n02, pp. 125--141,
  2002.

\bibitem{kolluri2004spectral}
R.~Kolluri, J.~R. Shewchuk, and J.~F. O'Brien, ``Spectral surface
  reconstruction from noisy point clouds,'' in \emph{Proceedings of the 2004
  Eurographics/ACM SIGGRAPH symposium on Geometry processing}.\hskip 1em plus
  0.5em minus 0.4em\relax ACM, 2004, pp. 11--21.

\bibitem{carr2001reconstruction}
J.~C. Carr, R.~K. Beatson, J.~B. Cherrie, T.~J. Mitchell, W.~R. Fright, B.~C.
  McCallum, and T.~R. Evans, ``Reconstruction and representation of 3d objects
  with radial basis functions,'' in \emph{Proceedings of the 28th annual
  conference on Computer graphics and interactive techniques}.\hskip 1em plus
  0.5em minus 0.4em\relax ACM, 2001, pp. 67--76.

\bibitem{kazhdan2013screened}
M.~Kazhdan and H.~Hoppe, ``Screened poisson surface reconstruction,'' \emph{ACM
  Transactions on Graphics (ToG)}, vol.~32, no.~3, p.~29, 2013.

\bibitem{hoppe1992surface}
H.~Hoppe, T.~DeRose, T.~Duchamp, J.~McDonald, and W.~Stuetzle, \emph{Surface
  reconstruction from unorganized points}.\hskip 1em plus 0.5em minus
  0.4em\relax ACM, 1992, vol.~26, no.~2.

\bibitem{boissonnat2002smooth}
J.-D. Boissonnat and F.~Cazals, ``Smooth surface reconstruction via natural
  neighbour interpolation of distance functions,'' \emph{Computational
  Geometry}, vol.~22, no. 1-3, pp. 185--203, 2002.

\bibitem{fleishman2005robust}
S.~Fleishman, D.~Cohen-Or, and C.~T. Silva, ``Robust moving least-squares
  fitting with sharp features,'' in \emph{ACM transactions on graphics (TOG)},
  vol.~24, no.~3.\hskip 1em plus 0.5em minus 0.4em\relax ACM, 2005, pp.
  544--552.

\bibitem{cheng2008survey}
Z.-Q. Cheng, Y.-Z. Wang, B.~Li, K.~Xu, G.~Dang, and S.-Y. Jin, ``A survey of
  methods for moving least squares surfaces.'' in \emph{Volume graphics}, 2008,
  pp. 9--23.

\bibitem{kazhdan2005reconstruction}
M.~Kazhdan, ``Reconstruction of solid models from oriented point sets,'' in
  \emph{Proceedings of the third Eurographics symposium on Geometry
  processing}.\hskip 1em plus 0.5em minus 0.4em\relax Eurographics Association,
  2005, p.~73.

\bibitem{kazhdan2006poisson}
M.~Kazhdan, M.~Bolitho, and H.~Hoppe, ``Poisson surface reconstruction,'' in
  \emph{Proceedings of the fourth Eurographics symposium on Geometry
  processing}, vol.~7, 2006.

\bibitem{qi2017pointnet}
C.~R. Qi, H.~Su, K.~Mo, and L.~J. Guibas, ``Pointnet: Deep learning on point
  sets for 3d classification and segmentation,'' in \emph{Proceedings of the
  IEEE Conference on Computer Vision and Pattern Recognition}, 2017, pp.
  652--660.

\bibitem{chen2018deep}
W.~Chen, X.~Han, G.~Li, C.~Chen, J.~Xing, Y.~Zhao, and H.~Li, ``Deep rbfnet:
  Point cloud feature learning using radial basis functions,'' \emph{arXiv
  preprint arXiv:1812.04302}, 2018.

\bibitem{broomhead1988radial}
D.~S. Broomhead and D.~Lowe, ``Radial basis functions, multi-variable
  functional interpolation and adaptive networks,'' Royal Signals and Radar
  Establishment Malvern (United Kingdom), Tech. Rep., 1988.

\bibitem{ben20183dmfv}
Y.~Ben-Shabat, M.~Lindenbaum, and A.~Fischer, ``3dmfv: Three-dimensional point
  cloud classification in real-time using convolutional neural networks,''
  \emph{IEEE Robotics and Automation Letters}, vol.~3, no.~4, pp. 3145--3152,
  2018.

\bibitem{sanchez2013image}
J.~S{\'a}nchez, F.~Perronnin, T.~Mensink, and J.~Verbeek, ``Image
  classification with the fisher vector: Theory and practice,''
  \emph{International journal of computer vision}, vol. 105, no.~3, pp.
  222--245, 2013.

\bibitem{li2018pointcnn}
Y.~Li, R.~Bu, M.~Sun, W.~Wu, X.~Di, and B.~Chen, ``Pointcnn: Convolution on
  x-transformed points,'' in \emph{Advances in Neural Information Processing
  Systems}, 2018, pp. 820--830.

\bibitem{wang2018dynamic}
Y.~Wang, Y.~Sun, Z.~Liu, S.~E. Sarma, M.~M. Bronstein, and J.~M. Solomon,
  ``Dynamic graph cnn for learning on point clouds,'' \emph{arXiv preprint
  arXiv:1801.07829}, 2018.

\bibitem{shen2018mining}
Y.~Shen, C.~Feng, Y.~Yang, and D.~Tian, ``Mining point cloud local structures
  by kernel correlation and graph pooling,'' in \emph{Proceedings of the IEEE
  conference on computer vision and pattern recognition}, 2018, pp. 4548--4557.

\bibitem{hua2018pointwise}
B.-S. Hua, M.-K. Tran, and S.-K. Yeung, ``Pointwise convolutional neural
  networks,'' in \emph{Proceedings of the IEEE Conference on Computer Vision
  and Pattern Recognition}, 2018, pp. 984--993.

\bibitem{liu2019relation}
Y.~Liu, B.~Fan, S.~Xiang, and C.~Pan, ``Relation-shape convolutional neural
  network for point cloud analysis,'' in \emph{Proceedings of the IEEE
  Conference on Computer Vision and Pattern Recognition}, 2019, pp. 8895--8904.

\bibitem{klokov2017escape}
R.~Klokov and V.~Lempitsky, ``Escape from cells: Deep kd-networks for the
  recognition of 3d point cloud models,'' in \emph{Proceedings of the IEEE
  International Conference on Computer Vision}, 2017, pp. 863--872.

\bibitem{li2018so}
J.~Li, B.~M. Chen, and G.~Hee~Lee, ``So-net: Self-organizing network for point
  cloud analysis,'' in \emph{Proceedings of the IEEE conference on computer
  vision and pattern recognition}, 2018, pp. 9397--9406.

\bibitem{lei2019octree}
H.~Lei, N.~Akhtar, and A.~Mian, ``Octree guided cnn with spherical kernels for
  3d point clouds,'' in \emph{Proceedings of the IEEE Conference on Computer
  Vision and Pattern Recognition}, 2019, pp. 9631--9640.

\bibitem{wu2019pointconv}
W.~Wu, Z.~Qi, and L.~Fuxin, ``Pointconv: Deep convolutional networks on 3d
  point clouds,'' in \emph{Proceedings of the IEEE Conference on Computer
  Vision and Pattern Recognition}, 2019, pp. 9621--9630.

\bibitem{fan2017point}
H.~Fan, H.~Su, and L.~J. Guibas, ``A point set generation network for 3d object
  reconstruction from a single image,'' in \emph{Proceedings of the IEEE
  conference on computer vision and pattern recognition}, 2017, pp. 605--613.

\bibitem{achlioptas2017learning}
P.~Achlioptas, O.~Diamanti, I.~Mitliagkas, and L.~Guibas, ``Learning
  representations and generative models for 3d point clouds,'' \emph{arXiv
  preprint arXiv:1707.02392}, 2017.

\bibitem{goodfellow2014generative}
I.~Goodfellow, J.~Pouget-Abadie, M.~Mirza, B.~Xu, D.~Warde-Farley, S.~Ozair,
  A.~Courville, and Y.~Bengio, ``Generative adversarial nets,'' in
  \emph{Advances in neural information processing systems}, 2014, pp.
  2672--2680.

\bibitem{wang2018pixel2mesh}
N.~Wang, Y.~Zhang, Z.~Li, Y.~Fu, W.~Liu, and Y.-G. Jiang, ``Pixel2mesh:
  Generating 3d mesh models from single rgb images,'' in \emph{Proceedings of
  the European Conference on Computer Vision (ECCV)}, 2018, pp. 52--67.

\bibitem{yang2018foldingnet}
Y.~Yang, C.~Feng, Y.~Shen, and D.~Tian, ``Foldingnet: Point cloud auto-encoder
  via deep grid deformation,'' in \emph{Proceedings of the IEEE Conference on
  Computer Vision and Pattern Recognition}, 2018, pp. 206--215.

\bibitem{yifan2019patch}
W.~Yifan, S.~Wu, H.~Huang, D.~Cohen-Or, and O.~Sorkine-Hornung, ``Patch-based
  progressive 3d point set upsampling,'' in \emph{Proceedings of the IEEE
  Conference on Computer Vision and Pattern Recognition}, 2019, pp. 5958--5967.

\bibitem{yu2018pu}
L.~Yu, X.~Li, C.-W. Fu, D.~Cohen-Or, and P.-A. Heng, ``Pu-net: Point cloud
  upsampling network,'' in \emph{Proceedings of the IEEE Conference on Computer
  Vision and Pattern Recognition}, 2018, pp. 2790--2799.

\bibitem{yu2018ec}
------, ``Ec-net: an edge-aware point set consolidation network,'' in
  \emph{Proceedings of the European Conference on Computer Vision (ECCV)},
  2018, pp. 386--402.

\bibitem{roveri2018pointpronets}
R.~Roveri, A.~C. {\"O}ztireli, I.~Pandele, and M.~Gross, ``Pointpronets:
  Consolidation of point clouds with convolutional neural networks,'' in
  \emph{Computer Graphics Forum}, vol.~37, no.~2.\hskip 1em plus 0.5em minus
  0.4em\relax Wiley Online Library, 2018, pp. 87--99.

\bibitem{huang2009consolidation}
H.~Huang, D.~Li, H.~Zhang, U.~Ascher, and D.~Cohen-Or, ``Consolidation of
  unorganized point clouds for surface reconstruction,'' \emph{ACM transactions
  on graphics (TOG)}, vol.~28, no.~5, p. 176, 2009.

\bibitem{park2019deepsdf}
J.~J. Park, P.~Florence, J.~Straub, R.~Newcombe, and S.~Lovegrove, ``Deepsdf:
  Learning continuous signed distance functions for shape representation,'' in
  \emph{Proceedings of the IEEE Conference on Computer Vision and Pattern
  Recognition}, 2019, pp. 165--174.

\bibitem{atzmon2019controlling}
M.~Atzmon, N.~Haim, L.~Yariv, O.~Israelov, H.~Maron, and Y.~Lipman,
  ``Controlling neural level sets,'' \emph{arXiv preprint arXiv:1905.11911},
  2019.

\bibitem{stutz2018learning}
D.~Stutz and A.~Geiger, ``Learning 3d shape completion from laser scan data
  with weak supervision,'' in \emph{Proceedings of the IEEE Conference on
  Computer Vision and Pattern Recognition}, 2018, pp. 1955--1964.

\bibitem{ronneberger2015u}
O.~Ronneberger, P.~Fischer, and T.~Brox, ``U-net: Convolutional networks for
  biomedical image segmentation,'' in \emph{International Conference on Medical
  image computing and computer-assisted intervention}.\hskip 1em plus 0.5em
  minus 0.4em\relax Springer, 2015, pp. 234--241.

\bibitem{bernardini1999ball}
F.~Bernardini, J.~Mittleman, H.~Rushmeier, C.~Silva, and G.~Taubin, ``The
  ball-pivoting algorithm for surface reconstruction,'' \emph{IEEE transactions
  on visualization and computer graphics}, vol.~5, no.~4, pp. 349--359, 1999.

\bibitem{tatarchenko2017octree}
M.~Tatarchenko, A.~Dosovitskiy, and T.~Brox, ``Octree generating networks:
  Efficient convolutional architectures for high-resolution 3d outputs,'' in
  \emph{Proceedings of the IEEE International Conference on Computer Vision},
  2017, pp. 2088--2096.

\bibitem{hane2017hierarchical}
C.~H{\"a}ne, S.~Tulsiani, and J.~Malik, ``Hierarchical surface prediction for
  3d object reconstruction,'' in \emph{2017 International Conference on 3D
  Vision (3DV)}.\hskip 1em plus 0.5em minus 0.4em\relax IEEE, 2017, pp.
  412--420.

\bibitem{dumoulin2016adversarially}
V.~Dumoulin, I.~Belghazi, B.~Poole, O.~Mastropietro, A.~Lamb, M.~Arjovsky, and
  A.~Courville, ``Adversarially learned inference,'' \emph{arXiv preprint
  arXiv:1606.00704}, 2016.

\end{thebibliography}

\end{document}